\documentclass[final,2p,times,twocolumn]{elsarticle}

\usepackage{lineno,hyperref}
\modulolinenumbers[1]

\journal{Swarm and Evolutionary Computation}

\usepackage{amsmath}
\usepackage{algorithm}
\usepackage[noend]{algpseudocode}
\usepackage{verbatim}
\usepackage[inline]{enumitem}
\usepackage{soul}

\usepackage{placeins} 

\usepackage{booktabs}
\usepackage{amssymb}
\usepackage{makecell}
\usepackage{multirow}
\usepackage{graphicx}
\usepackage{rotating}

\usepackage{colortbl}

\definecolor{Gray}{gray}{0.90}
\definecolor{LightCyan}{rgb}{0.88,1,1}

\newcolumntype{a}{>{\columncolor{LightCyan}}c}
\newcolumntype{b}{>{\columncolor{white}}c}

\usepackage{tikz}
\usetikzlibrary{shapes.geometric, arrows}
\tikzstyle{squarered} = [rectangle, minimum width=3cm, minimum height=1cm, text centered, draw=black, fill=red!10]
\tikzstyle{squaregreen} = [rectangle, minimum width=3cm, minimum height=1cm, text centered, draw=black, fill=green!10]
\tikzstyle{squareblue} = [rectangle, minimum width=3cm, minimum height=1cm, text centered, draw=black, fill=blue!10]
\tikzstyle{diamondgreen} = [diamond, minimum width=3cm, minimum height=1cm, text centered, draw=black, fill=green!10]
\tikzstyle{diamondblue} = [diamond, minimum width=3cm, minimum height=1cm, text centered, draw=black, fill=blue!10]
\tikzstyle{diamondred} = [diamond, minimum width=3cm, minimum height=1cm, text centered, draw=black, fill=red!10]
\tikzstyle{diamondorange} = [diamond, minimum width=3cm, minimum height=1cm, text centered, draw=black, fill=orange!10]
\tikzstyle{circlewhite} = [circle, minimum width=0.5cm, minimum height=0.5cm, text centered, draw=black]
\tikzstyle{circleyellow} = [circle, minimum width=1cm, minimum height=1cm, text centered, draw=black, fill=yellow!10]
\tikzstyle{squareryellow} = [rectangle, rounded corners, minimum width=3cm, minimum height=1cm, text centered, draw=black, fill=yellow!10]
\tikzstyle{invisible} = [rectangle, minimum width=0.5cm, minimum height=0.5cm, text centered]
\tikzstyle{arrow} = [thick,->,>=stealth]
\tikzstyle{darrow} = [dashed,->,>=stealth]
\tikzstyle{line} = [thick,-,>=stealth]

\newlist{inparaenum}{enumerate*}{1}
\setlist[inparaenum]{label=(\roman*)}

\def\CC{{C\nolinebreak[4]\hspace{-.05em}\raisebox{.4ex}{\tiny\bf ++}}}

\usepackage{xcolor,colortbl}

\usepackage[english]{babel}

\algrenewcommand\alglinenumber[1]{{\sffamily\footnotesize#1}}

\begin{document}

\begin{frontmatter}

\title{On Explaining Machine Learning Models \\by Evolving Crucial and Compact Features}

\author{Marco Virgolin}
\address{Life Sciences and Health Group, Centrum Wiskunde \& Informatica, Amsterdam 1098 XG, the Netherlands}

\author{Tanja Alderliesten}
\address{Department of Radiation Oncology, Amsterdam UMC, Amsterdam 1105 AZ, the Netherlands}

\author{Peter A.N. Bosman}
\address{Life Sciences and Health Group, Centrum Wiskunde \& Informatica, Amsterdam 1098XG, the Netherlands}
\address{Algorithmics Group, Delft University of Technology, Delft 2628 XE, the Netherlands}

\begin{abstract}
Feature construction can substantially improve the accuracy of Machine Learning (ML) algorithms. Genetic Programming (GP) has been proven to be effective at this task by evolving non-linear combinations of input features. GP additionally has the potential to improve ML explainability since explicit expressions are evolved. Yet, in most GP works the complexity of evolved features is not explicitly bound or minimized though this is arguably key for explainability. In this article, we assess to what extent GP still performs favorably at feature construction when constructing features that are (1) Of small-enough number, to enable visualization of the behavior of the ML model; (2) Of small-enough size, to enable interpretability of the features themselves; (3) Of sufficient informative power, to retain or even improve the performance of the ML
algorithm. We consider a simple feature construction scheme using three different GP algorithms, as well as random search, to evolve features for five ML algorithms, including support vector machines and random forest. Our results on 21 datasets pertaining to classification and regression problems show that constructing only two compact features can be sufficient to rival the use of the entire original feature set. We further find that a modern GP algorithm, GP-GOMEA, performs best overall. These results, combined with examples that we provide of readable constructed features and of 2D visualizations of ML behavior, lead us to positively conclude that GP-based feature construction still works well when explicitly searching for compact features, making it extremely helpful to explain ML models.\\ \\
\scriptsize{
This preprint is associated to a manuscript accepted for publication on \emph{Swarm and Evolutionary Computation}, \texttt{doi.org/10.1016/j.swevo.2019.100640}. \\ This work is licensed under a Creative Commons ``Attribution-NonCommercial-NoDerivs 3.0 Unported'' license.
}
\end{abstract}


\end{frontmatter}

\section{Introduction}

Feature selection and feature construction are two important steps to improve the performance of any Machine Learning (ML) algorithm~\cite{liu1998feature,friedman2001elements}. Feature selection is the task of excluding features that are redundant or misleading. Feature construction is the task of transforming (parts of) the original feature space into one that the ML algorithm can better exploit.

A very interesting method to perform feature construction automatically is Genetic Programming (GP)~\cite{koza1992gp,poli2008field}.
GP can synthesize functions without many prior assumptions on their form, differently from, e.g., logistic regression or regression splines \cite{friedman1991multivariate,hosmer2013applied}. Moreover, feature construction not only depends on the data at hand, but also on the way a specific ML algorithm can model that data. Evolutionary methods in general are highly flexible in their use due to the way they perform search (i.e., derivative free). This makes it possible, for example, to evaluate the quality of a feature for a specific ML algorithm by directly measuring what its impact is on the performance of the ML algorithm (i.e., by training and validating the ML algorithm when using that feature).

\begin{figure}
\centering
\includegraphics[width=0.99\linewidth]{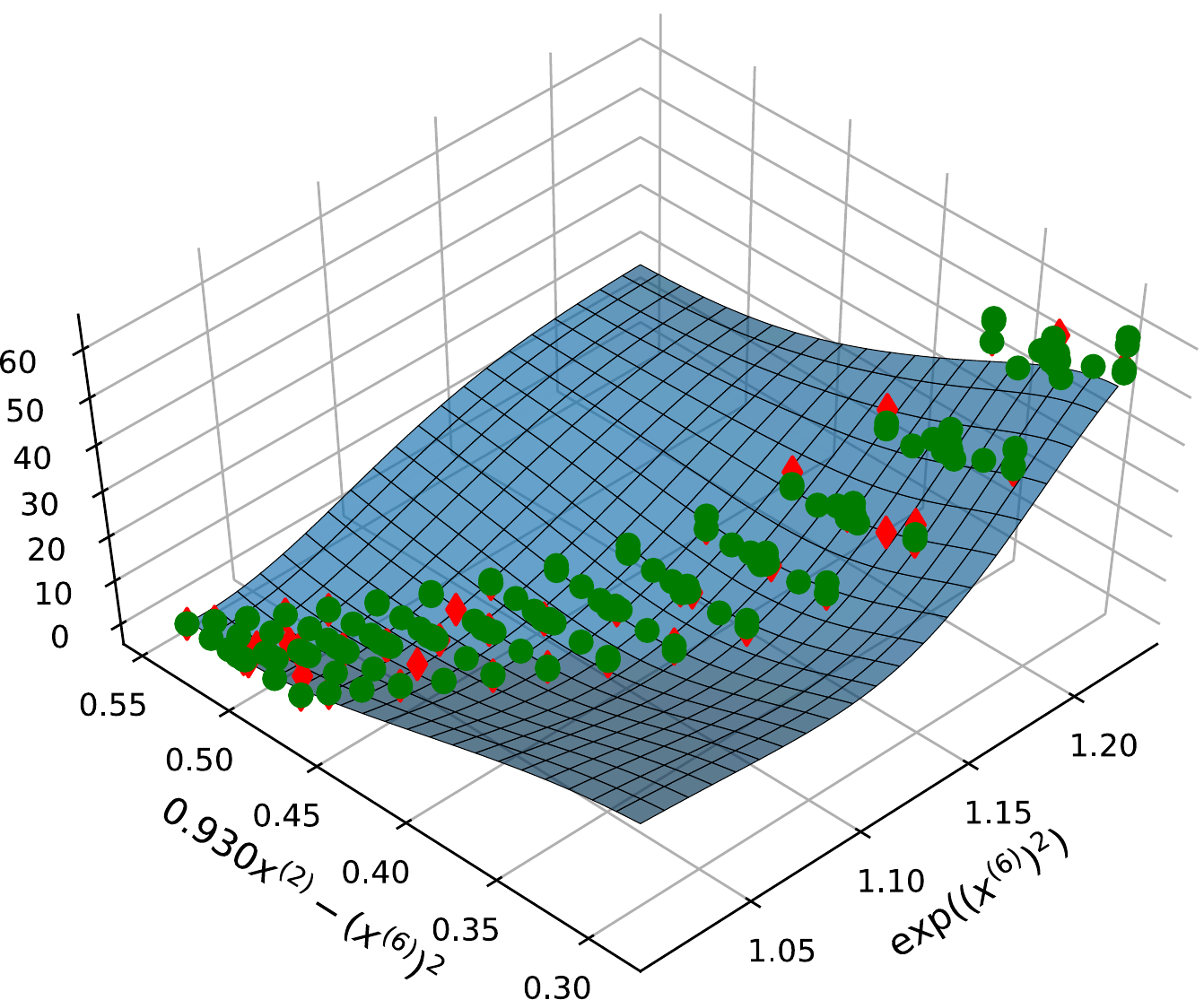}
\caption{Regression surface learned by SVM for the Yacht dataset (in blue), expressed as a 2D function of the two features (on the bottom axes) constructed by our approach. Circles are training samples, diamonds are test samples.
The dataset has six features ($x^{(i)}$). Our approach constructs two new features (using GP-GOMEA, see Sec.~\ref{sec:details-gp}), which are non-linear transformations of the prismatic coefficient ($x^{(2)}$) and the Froude number ($x^{(6)}$). With only two features the SVM prediction surface can be visualized. Moreover, these new features are understandable. Finally, the modeling quality is actually improved over employing SVM directly on all six features. The coefficient of determination of SVM increased from 85\% using the original features to 98\% using the two new features.}\label{fig:svm-yacht}
\vspace{-3mm}
\end{figure}

Explaining what constructed features mean can shed light on the behavior of ML-inferred models that use such features. Reducing the number of features is also important to improve interpretability. If the original feature space is reduced to few constructed features (e.g., up to two for regression and up to three for classification), the function learned by the ML model can be straightforwardly visualized w.r.t. the new features. 
In fact, how to make ML models more understandable is a key topic of modern ML research, as many practical, sensitive applications exist, where explaining (part of) the behavior of ML models is essential to trust their use (e.g., in medical applications)~\cite{lipton2018mythos,guidotti2018survey,adadi2018peeking,goodman2017european}.
Typically, GP for feature construction searches in a subspace of mathematical expressions. Adding to the appeal and potential of GP, these expressions can be human-interpretable if simple enough~\cite{guidotti2018survey,virgolin2019model}. 

Figure~\ref{fig:svm-yacht} presents an example of the potential held by such an approach: a multi-dimensional dataset transformed into a 2D one, where both the behavior of the ML algorithm and the meaning of the new features is clear, while the performance of the ML algorithm is not compromised w.r.t. the use of the original feature set (it is actually improved).

In this article we study whether GP can be useful to construct a \emph{low} number of \emph{small} features, to increase the chance of obtaining interpretable ML models, without compromising their accuracy (compared to using the original feature set). To this end, we design a simple, iterative feature construction scheme, and perform a wide set of experiments: we consider four types of feature construction methods (three GP algorithms and random search), five types of machine learning algorithms. We apply their combinations on 21 datasets between classification and regression to determine to what extent they are capable of effectively and efficiently finding crucial and compact features for specific ML algorithms.

The main original scientific contribution of this work is an investigation of whether GP can be used to construct features that are:
\begin{itemize}
	\item Of small-enough number, to enable visualization of the behavior of the ML model;
	\item Of small-enough size, to enable interpretability of the features themselves;
	\item Of sufficient informative power, to retain or even improve the performance of the ML, compared to using the original feature set;
\end{itemize}
These aspects are assessed under different circumstances:
\begin{itemize}
	\item We test different search algorithms, including modern model-based GP and random search;
	\item We test different ML algorithms.
\end{itemize}

The remainder of this article is organized as follows. Related work is reported in Section~\ref{sec:related}. The proposed feature construction scheme is presented in Section~\ref{sec:scheme}. The search algorithms to construct features, as well as the considered ML algorithms, are presented in Section~\ref{sec:algos}. The experimental setup is described in Section~\ref{sec:experiments}. Results related to performance are reported in Section~\ref{sec:result-rq12}, while results concerning interpretability are reported in Section~\ref{sec:result-rq3}. Section~\ref{sec:discussion} discusses our findings, and Section~\ref{sec:conclusion} concludes this article.

\section{Related work}\label{sec:related}
In this article, we consider GP for feature construction to achieve better explainable ML models. Different forms of GP to obtain explainable ML have been explored in literature, but they do not necessarily leverage feature construction. E.g.,~\cite{cano2013interpretable} introduced a form of GP for the automatic synthesis of interpretable classifiers, generated from scratch as self-contained ML models, made of IF-THEN rules.
A very different paradigm for explainable ML by GP is considered in~\cite{evans2019s}, where the authors explore the use of GP to recover the behavior of a given unintelligible classifier by evolving interpretable approximation models.
Other GP-based approaches and paradigms to synthesize interpretable ML models from scratch, or to approximate the behavior of pre-existing ML models by interpretable expressions, are reported in recent surveys on explainable artificial ingelligence such as~\cite{guidotti2018survey,adadi2018peeking}.

Since in this article we particularly study what the potential of GP for feature construction is in terms of added value for explaining complex, not directly explainable models learned by various popular ML algorithms, the related work that follows describes GP approaches for feature construction. For readers interested in feature selection, we refer to a recent survey~\cite{xue2016survey}.

One of the first approaches of GP for feature construction is presented in~\cite{krawiec2002genetic}. There, each GP solution is a set of $K$ features. The fitness of a set is the cross-validation performance of a decision tree~\cite{breiman2017classification} using that set. The results on six classification datasets show that the approach is able to synthesize a feature set that is competitive with the original one, and can also be added to the original set for further improvements.
No attention is however given to the interpretability of evolved features.

The work in~\cite{muharram2005evolutionary} generates one feature with Standard, tree-based GP (SGP)~\cite{koza1992gp}, to be added to the original set. Feature importance metrics of decision trees such as information gain, Gini index and $Chi^2$ are used as fitness measure. An advantage of using such fitness measures over ML performance is that they can be computed very quickly. However, they are decision tree-specific. Results show that the approach can improve prediction accuracy, and, for a few problems, it is shown that decision trees that are simple enough to be reasonably interpretable, can be found.
	
Feature construction for high-dimensional datasets is considered in~\cite{tran2016genetic}, for eight bio-medical binary classification problems, with 2,000 to 24,188 features. This approach is different from the typical ones, as the authors propose to use SGP to evolve classifiers rather than features, and extract features from the components (subtrees) of such classifiers. These are then used as new features for an ML algorithm. Results on K-Nearest Neighbors~\cite{altman1992introduction}, Naive Bayes classifier~\cite{russell2016artificial,murphy2006naive}, and decision tree show that a so-found feature set can be competitive or outperform the original one. The authors show an example where a single interpretable feature is constructed that enables linear separation of the classification examples. 

Different from the aforementioned works, \cite{chen2017genetic} explores feature construction for regression. A SGP-based approach is designed to tackle regression problems with a large number of features, and is tested on six datasets. Instead of using the constructed features for a different ML algorithm, SGP dynamically incorporates them within an ongoing run, to enrich the terminal set. Every $\alpha$ generations of SGP, the subtrees composing the best solutions become new features by encapsulation into new terminal nodes. The approach is found to improve the ability of SGP to find accurate solutions. However, the features found by encapsulating subtrees are not interpretable because allowing subsequent encapsulations leads to an exponential growth of solution size.

A recent work that focuses on evolutionary dimensionality reduction and consequent visualization is~\cite{cano2017multiobjective}, where a multi-objective, grammar-based SGP approach is employed. $K$ feature transformations are evolved in synergy to enable, at the same time, good classification accuracy, and visualization through dimensionality reduction. The system is thoroughly tested on 42 classification tasks, showing that the algorithm performs well compared to state-of-the-art dimensionality reduction methods, and it enables visualization of the learned space. However, as trees are free to grow up to a height of 50, the constructed features themselves cannot be interpreted. 

The most similar works to ours that we found are~\cite{virgolin2018gecco} and~\cite{tran2019genetic}. In~\cite{virgolin2018gecco}, which is our previous work, the possibility of using a modern model-based GP algorithm (which we also use in our comparisons) for feature construction is explored on four regression datasets. There, focus is put on keeping feature size small, to actively attempt to obtain readable features. These features are iteratively constructed to be added to the original feature set to improve the performance of the ML algorithm, and three ML algorithms are compared (linear regression, support vector machines~\cite{cortes1995support}, random forest~\cite{breiman2001random}). Reducing the feature space to enable a better understanding of inferred ML models is not considered.

In~\cite{tran2019genetic}, different feature construction approaches are compared on gene-expression datasets that have a large number of features (thousands to tens of thousands) to study if evolving class-dependent features, i.e., features that are each targeted at aiding the ML algorithm detect one specific class, can be beneficial. Similarly to us, the authors show visualizations of feature space reduced to up to three constructed features, and an example of three features that are encoded as very small, easy-to-interpret trees. However, such small features are a rare outcome as the trees used to encode features typically had more than 75 nodes. These trees are therefore arguably extremely hard to read and interpret.

Our work is different from previous research in two major aspects. First, none of the previous work principally addresses the conflicting objectives of retaining good performance of an ML algorithm while attempting to explain both its behavior (by dimensionality reduction to allow visualization), and the meaning of the features themselves (by constraining feature complexity).
Second, multiple GP algorithms within a same feature construction scheme, on multiple ML algorithms, are not compared in previous work. Most of the times, it is a different feature construction scheme that is tested, using arguably small variations of SGP. Here, we consider random search, two versions of SGP, as well as another modern GP algorithm. Furthermore, we adopt both ``weak'' ML algorithms such as ordinary least squares linear regression and the naive Bayes classifier, as well as ``strong'', state-of-the-art ones, which are rarely used in literature for feature construction, such as support vector machine and  random forest; on both classification and regression tasks.

\begin{figure*}[h]

\centering
\scalebox{0.7}{
\begin{tikzpicture}[node distance=4cm]
\centering

\node (Training) [squaregreen] {
\scalebox{0.90}{
\begin{tabular}{cccc}
\multicolumn{4}{c}{$Tr$ iteration $k-1$}\\
$x^{(1)}$ & \dots & $x^{(k-1)} $ & $y$ \\
\midrule
22.49 & \dots & -3.10 & 10.4\\
12.98 & \dots & -7.41 & 7.49\\
\dots & \dots & \dots & \dots \\
\end{tabular}
}
};

\node (GP) [diamondorange, right of=Training, xshift=1cm, yshift=2.0cm] {GP};

\node (eTraining) [squaregreen, right of=GP, xshift=1.25cm] {
\scalebox{0.90}{
\begin{tabular}{c c >{\columncolor[gray]{0.8}}c c}

\multicolumn{4}{c}{$Tr$ iteration $k$}\\
\dots & $ x^{(k-1)} $ & $ x^{(k)} $ & $y$ \\
\midrule
\dots & -3.10 & 7.12 & 10.4\\
\dots & -7.41 & 9.41 & 7.49\\
\dots & \dots & \dots & \dots \\
\end{tabular}
}
};

\node (eTest) [squareblue, below of=eTraining, yshift=0.25cm] {
\scalebox{0.88}{
\begin{tabular}{c c >{\columncolor[gray]{0.8}}c c}
\multicolumn{4}{c}{$Te$ iteration $k$}\\
\dots & $ x^{(k-1)} $ & $ x^{(k)} $ & $y$ \\
\midrule
\dots & 9.87 & 1.11 & 5.55\\
\dots & 6.45 & 4.78 & 12.01\\
\dots & \dots & \dots & \dots \\
\end{tabular}
}
};

\node(eFeatureContainer) [invisible, below of=GP, xshift=-0.0cm, yshift=0cm, minimum height=4cm,minimum width=2.0cm]{};

\node(eFeatureN1) [circlewhite, below of=GP, yshift=1.75cm]{
};

\node(eFeatureN2) [circlewhite, below of=GP, xshift=-0.5cm, yshift=1.0cm]{
};

\node(eFeatureN3) [circlewhite, below of=GP, xshift=0.5cm, yshift=1.0cm]{
};

\node(eFeatureN4) [circlewhite, below of=GP, xshift=-1cm, yshift=0.25cm]{
};

\node(eFeatureN5) [circlewhite, below of=GP, xshift=0cm, yshift=0.25cm]{
};

\node(eFeatureCap) [invisible, below of=GP, yshift=-0.5cm] {
	Best New Feature $x^{(k)}$
};

\node(ML algorithm) [diamondorange, right of=eTraining, xshift=1.5cm] {
ML alg.
};

\node(Model) [squarered, below of=ML algorithm, yshift=1cm] {
 Trained Model
};

\node(Prediction) [invisible, below of=Model, yshift=1.5cm] {
 Test Error $k$
};

\draw [darrow] (Training) |- (GP);

\draw [arrow] (GP) -- (eFeatureContainer);
\draw [arrow] (eFeatureContainer) ++(25pt,25pt) -| ++(20pt,40pt) |- ++(46pt,25pt);
\draw [arrow] (eFeatureContainer) ++(25pt,25pt) -| ++(20pt,40pt) |- (eTest);
\draw [line] (eFeatureN1) -- (eFeatureN2);
\draw [line] (eFeatureN1) -- (eFeatureN3);
\draw [line] (eFeatureN2) -- (eFeatureN4);
\draw [line] (eFeatureN2) -- (eFeatureN5);

\draw [darrow] (eTraining) -- (ML algorithm);
\draw [arrow] (ML algorithm) -- (Model);
\draw [darrow] (eTest) ++(63pt,21pt) -- (Model);
\draw [arrow] (Model) -- (Prediction);

\end{tikzpicture}
}
\caption{Construction of the $k$-th feature and computation of the $k$-th test error. Evolved features use the features of the original dataset (not shown) and random constants as terminal nodes. Dashed arrows represent inputs, solid arrows represents outputs.}\label{fig:featureevolutionscheme}
\end{figure*}
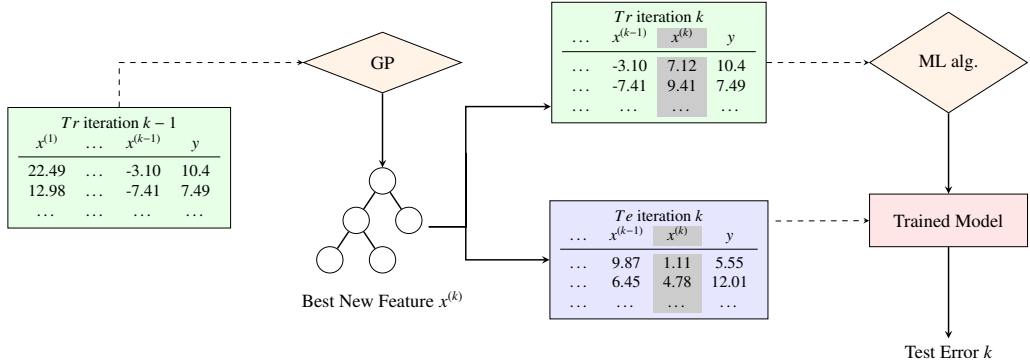

\section{Iterative evolutionary feature construction}\label{sec:scheme}
We use a remarkably simple scheme to construct features. 
Our approach constructs $K \in \mathbb{N}^+$ features by iterating $K$ GP runs. The evolution of the $k$-th feature ($k \in \{1, \dots, K\}$) uses the previously constructed $k-1$ features.

\subsection{Feature construction scheme}
The dataset $D$ defining the problem at hand is split into two parts: the training $Tr$ and the test $Te$ set. This partition is kept fixed through the whole procedure. Only $Tr$ is used to construct features, while $Te$ is exclusively used for final evaluation to avoid positive bias in the results~\cite{kohavi1997wrappers}. We use the notation $x^{(i)}_j$ to refer to the $i$-th feature value of the $j$-th example, and $y_j$ for the desired outcome (label for classification or target value for regression) of the $j$-th example.

The $k$-th GP run evolves the $k$-th feature. An example is shown in Figure~\ref{fig:featureevolutionscheme}.  Each solution in the population competes to become the new feature $x^{(k)}$, that represents a transformation of the original feature set. In every run, the population is initialized at random.

We evaluate the fitness of a feature of the $k$-th run by measuring the performance of the ML algorithm on a dataset that contains that feature and the previously evolved $k-1$ features.

We only use original features (and random constants) as terminals. In particular, the features constructed by previous iterations are not used as terminal nodes in the $k$-th run. This prevents the generation of nested features, which could harm interpretability.

At the end of the $k$-th run, the best feature is stored and its values $x^{(k)}_j$ are added to $Tr$ and $Te$ for the next iterations.

\subsection{Feature fitness}\label{sec:feature-fitness}
The fitness of a feature is computed by measuring the performance (i.e., error) of the ML algorithm when the new feature is added to $Tr$. We consider the $C$-fold cross-validation error rather than the training error to promote generalization and prevent overfitting. The pseudo code of the evaluation function is shown in Algorithm~\ref{alg:feature-fitness-computation}.

Specifically, the $C$-fold cross-validation error is computed by partitioning $Tr$ into $C$ splits. For each $c = 1, \dots, C$ iteration, a different split is used for validation (set $V^c$), and the remaining $C-1$ splits are used for training (set $Tr^c$).
The mean validation error is the final result.

For classification tasks, in order to take into account both multiple and possibly imbalanced class distributions, the prediction error is computed as 1 minus the \emph{macro} $F1$ score, i.e., 1 minus the mean of the class-specific $F1$ scores:

\begin{small}
\begin{align*}
1 - F1 & = 1 - \frac{1}{ \# \textit{classes} }  \sum_{ \gamma \in \textit{classes} }  F1_\gamma  
\\
& = 1 - \frac{2}{ \# \textit{classes} }  \sum_{ \gamma \in \textit{classes} }  \frac{  \frac{TP_\gamma}{TP_\gamma + FP_\gamma} \frac{TP_\gamma}{TP_\gamma + FN_\gamma}  }{ \frac{TP_\gamma}{TP_\gamma + FP_\gamma} + \frac{TP_\gamma}{TP_\gamma + FN_\gamma}  },
\end{align*} 
\end{small}
where $TP_\gamma, FN_\gamma, FP_\gamma$ are the true positive, false negative, and false positive classifications for the class $\gamma$, respectively. If the computation of $F1_\gamma$ results in $\frac{0}{0}$, we set $F1_\gamma = 0$.

For regression, the prediction error is computed with the Mean Squared Error (MSE).
 
\begin{algorithm}
\caption{Computation of the fitness of a feature $s$}\label{alg:feature-fitness-computation}
\begin{algorithmic}[1]
\scriptsize
\Function{ComputeFeatureFitness}{$s$}
\State $Tr^\prime \gets $AddFeatureToCurrentTrainingSet($s$)
\State $error \gets 0$
\For {$c = 1, \dots, C$}
	\State $T^c, V^c \gets $SplitSet($c,C,Tr^{\prime}$)
	\State $M \gets $TrainMLModel($T^c$)
	\State $error \gets error + $ComputeError($M, V^c$)
\EndFor
\State Return$\left( \frac{error}{C} \right) $
\EndFunction
\end{algorithmic}
\end{algorithm}

\subsection{Preventing unnecessary fitness computations}\label{sec:criteriaevaluation} 
Computing the fitness of a feature is particularly expensive, as it consists of a $C$-fold cross-validation of the ML algorithm. This limits the feasibility of, e.g., adopting large population sizes and large numbers of evaluations for the GP algorithms. 

We therefore attempt to prevent unnecessary cross-validation calls, by assessing if features meet four criteria. Let $n$ be the number of examples in $Tr$. The criteria are the following:
\begin{enumerate}
\item \emph{The feature is not a constant}.
We avoid evaluating constant features as they are likely to be useless for many ML algorithms, which internally already compute an intercept.
\\ \vspace{-.35cm}
\item \emph{The feature does not contain extreme values that may cause numerical errors}, i.e., with absolute value above a lower-bound $\beta_\ell$ or above an upper-bound $\beta_u$. Here, we set $\beta_\ell = 10^{-10}$, and $\beta_u = 10^{10}$ (none of the datasets considered here have values exceeding these bounds).
\\ \vspace{-.3cm}
\item \emph{The feature is not equivalent to one constructed in the previous $k-1$ iterations}. Equivalence is determined by checking the values available in $Tr$, i.e., equivalence holds if:
\begin{small}
\begin{equation*} 
\forall j \in Tr, \exists i \in \{1, \dots, k-1\} : x^{(k)}_j = x^{(i)}_{j}.
\end{equation*}
\end{small}
Note that a constructed feature that is equivalent to a feature of the original feature set can be valid, as long as no other previously constructed feature exists that is already equivalent. Thus, our approach can in principle perform pure feature selection.
\\ \vspace{-.35cm}
\item \emph{The values of the feature in consideration have changed since the last time the feature was evaluated}. GP variation can change the syntax of a feature without necessarily affecting its behavior (e.g., inserting a multiplication by 1 will not change the final values a feature computes). If the values do not change, then the fitness of the feature will not change either (see Sec.~\ref{sec:feature-fitness}).
We therefore avoid unnecessary re-computations of feature fitnesses, by caching the feature values prior to GP variation, and checking whether they have changed after variation.
\end{enumerate}

The computational effort for each criterion is $O(n)$ (it is $O((k-1)n)$ for criterion 3, however in our experiments $k \ll n$). The fitness of a feature failing criterion 1, 2, or 3 is set to the maximum possible error value. If criterion 4 fails, the fitness remains the same (although performing cross-validation may lead to slightly different results when using stochastic ML algorithms like random forest).

\section{Considered search algorithms and machine learning algorithms}\label{sec:algos}
We consider SGP, Random Search (RS), and the GP instance of the Gene-pool Optimal Mixing Evolutionary Algorithm (GP-GOMEA) as competing search algorithms to construct features. 
SGP is widely used in feature construction (see related work in Sec.~\ref{sec:related}). RS is not typically considered, yet we believe it is important to assess whether evolution does bring any benefit over random enumeration within the confines of our study, i.e., when forcing to find small features. GP-GOMEA is a recently introduced GP algorithm that has proven to be particularly proficient in evolving accurate solutions of limited size~\cite{virgolin2019model,virgolin2017scalable,virgolin2018gecco}.

As ML algorithms, we consider the Naive Bayes classifier (NB), ordinary least-squares Linear Regression (LR), Support Vector Machines (SVM), Random Forest (RF), and eXtreme Gradient Boosting (XGB). NB is used only for classification tasks, LR only for regression tasks, SVM, RF, and XGB for both tasks. We provide more details in the following sections.

\subsection{Details on the search algorithms}\label{sec:details-gp}
All search algorithms use the fitness evaluation function. A feature $s$ is evaluated by first checking whether the four criteria of Section~\ref{sec:criteriaevaluation} are met, and then, if the outcome is positive, by running the ML algorithm over the feature-extended dataset.

For SGP, we use subtree crossover and subtree mutation, picking the depth of subtree roots uniformly randomly as proposed in~\cite{pawlak2015semantic}. The candidate parents for variation are chosen with tournament selection.
Since we are interested in constructing small features so as to increase the chances they will be interpretable, we consider two versions of SGP. The first is the classic one where solutions are free to grow to tree heights typically much larger than the one used for tree initialization. In the following, the notation SGP refers to this first version. The second one uses trees that are not allowed to grow past the initial maximum tree height. We call this version \emph{bounded} SGP, and use the notation SGP\textsubscript{b}.

RS is realized by continuously sampling and evaluating new trees, keeping the best~\cite{koza1992gp}. Like for SGP\textsubscript{b}, a maximum tree height is fixed during the whole run. If evolution is hypothetically no better than RS, then we expect that SGPb and GP-GOMEA will construct features that are no better than the ones constructed by RS.

GP-GOMEA is a recently introduced GP algorithm that has been found to deliver accurate solutions of small size on benchmark problems~\cite{virgolin2017scalable}, and to work well when a small size is enforced in symbolic regression~\cite{virgolin2019model,virgolin2018gecco}.
GP-GOMEA uses a tree template fixed by a maximum tree height (which can include intron nodes to allow for unbalanced tree shapes) and performs homologous variation, i.e., mixed tree nodes come from the same positions in the tree. Each generation prior to mixing, a hierarchical model that captures interdependencies (\emph{linkage}) between nodes is built (using mutual information). This model, called Linkage Tree (LT), drives variation by indicating what nodes should be changed \emph{en block} during mixing, to avoid the disruption of patterns with large linkage.
 
The LT has been shown to enable GP-GOMEA to outperform subtree crossover and subtree mutation of SGP, as well as the use of a randomly-build LT, i.e., the Random Tree (RT), on problems of different nature~\cite{virgolin2019model,virgolin2017scalable}. 
However, the LT requires sufficiently large population sizes to be accurate and beneficial (e.g., several thousand solutions in GP for symbolic regression)~\cite{virgolin2019model}. Because in the framework of this article fitness evaluations use the cross-validation of a ML algorithm, we cannot afford to use large population sizes. Accordingly, we found the adoption of the LT to not be superior to the adoption of the RT under these circumstances in preliminary experiments. Therefore, for the most part, we adopt GP-GOMEA with the RT (GP-GOMEA\textsubscript{RT}). This means we effectively compare random hierarchical homologous variation with subtree-based variation. An example of adopting the LT and large population sizes for feature construction is provided in Section~\ref{sec:discussion}.

\subsection{Details on the ML algorithms}\label{sec:details-ML algorithms}
We now briefly describe the ML algorithms used in this work: NB, LR, SVM, RF, and XGB. NB and LR are less computationally expensive compared to SVM, RF, and XGB.
Details on the computational time complexity of these algorithms are reported at: https://bit.ly/2PG0xse. 

NB is a classifier which assumes independence between features~\cite{russell2016artificial,murphy2006naive}. NB is often used as a baseline, as it is simple and fast to train. 
We use the \emph{mlpack} implementation of NB~\cite{mlpack2013} and assume the data to be normally distributed (default setting).

Similarly to NB, LR is often used as a baseline as it is simple and fast, for regression tasks. LR assumes that the target variable can be explained by a linear combination of the features~\cite{russell2016artificial}. 
We use the \emph{mlpack} implementation of LR~\cite{mlpack2013}.

SVM is a powerful ML algorithm that can be used for non-linear classification and regression~\cite{cortes1995support,CC01a}. 
We use the \emph{libsvm} \CC~implementation~\cite{CC01a}. 
We consider the Radial Basis Function (RBF) kernel, which works well in practice for many problems, with C-SVM for classification, and $\mathcal{E}$-SVM for regression.

RF is an ensemble ML algorithm which, like SVM, can be used for both classification and regression and can infer non-linear patterns~\cite{breiman2001random}. 
RF builds an ensemble of (typically deep) decision trees, each trained on a sample of the training set (\emph{bagging}). At prediction time, the mean (or maximum agreement) prediction of the decision trees is returned. 
We use the \emph{ranger} \CC~implementation~\cite{wright2015ranger}.

XGB is, like RF, an ensemble ML algorithm, typically based on decision trees, and capable of learning non-linear models~\cite{chen2016xgboost}. XGB works by boosting, i.e., stacking together multiple weak estimators (small decision tress) that fit the data in an incremental fashion.
We use the \emph{dmlc} \CC~implementation (https://bit.ly/34fBNeA).

\begin{table}
\caption{Parameter settings of the GP algorithms.}
\label{tab:ea-parameters}
\small
\centering
\scalebox{0.80}{
\begin{tabular}{lcc}
\toprule
 & SGP(b) & GP-GOMEA\textsubscript{RT} \\
\midrule
Population size & $100$ & $100$  \\
Initialization method & Ramped Half and Half & Half and Half  \\
Initialization max tree height & $2\text{--}6$ ($2$ or $4$) & $2$ or $4$ \\
Max tree height & 17 ($2$ or $4$) & $2$ or $4$ \\
Variation & SX $0.9$, SM $0.1$ & parameter-less \\
Selection & \makecell{Tournament 7, Elitism 1} & parameter-less \\
\hline
Function set & \multicolumn{2}{c}{ $\{ +, \times, -, \div, \cdot^2, \sqrt{\cdot}, \log_p, \exp \}$ for all} \\
Terminal set & \multicolumn{2}{c} { $\{ x^{(i)}, \texttt{ERC} \}$ for all} \\
\bottomrule
\end{tabular}
}
\end{table}

\section{Experiments}\label{sec:experiments}
We perform 30 runs of our Feature Construction Scheme (FCS), with SGP, SGP\textsubscript{b}, RS, and GP-GOMEA\textsubscript{RT}, in combination with each ML algorithm (NB only for classification and LR only for regression), on each problem. 
Each run of the FCS uses a random train-test split of 80\%-20\%, and considers up to $K=5$ features construction rounds.
We use a population size of 100 for the search algorithms, and assign a maximum budget of $10,000$ function evaluations to each FCS iteration. This results in relatively large running times for complex ML algorithms (see Sec.~\ref{sec:running-time}).
An experiment including larger evolutionary budgets and the use of the LT in GP-GOMEA is presented in the discussion (Sec.~\ref{sec:discussion}).
We use a limit on the total number of evaluations instead of a a limit on the total number of generations because GP-GOMEA\textsubscript{RT} performs more evaluations than SGP per generation~\cite{virgolin2017scalable}.

For GP-GOMEA\textsubscript{RT}, SGP\textsubscript{b}, and RS, we consider two levels of maximum tree height $h$: 2 and 4. This choice yields a maximum solution size of 7 and 31 respectively (using function nodes with a maximum arity $r=2$). We choose these two height levels because we found features with $h=2$ to be arguably easy to read and interpret, whereas features with $h=4$ can already be very hard to understand. This indication is also reported in~\cite{virgolin2019model} for the evolution of symbolic regression formulas. Note that using a tree height limit over a solution size limit prevents finding deep trees containing the nesting of the arguably more complicated to understand non-linear functions $\cdot^2, \sqrt{\cdot}, \log_p, \exp$.
We do not consider bigger tree heights as resulting features may likely be impossible to interpret, defying a key focus of this work.

Other parameter settings used for the GP algorithms are shown in Table~\ref{tab:ea-parameters}. SGP\textsubscript{b} uses the same settings as SGP, except for the maximum tree height (at initialization and along the whole run), which is set to the same of GP-GOMEA\textsubscript{RT}. In GP-GOMEA\textsubscript{RT} we use the Half and Half (HH) tree initialization method instead of the Ramped Half and Half (RHH)~\cite{koza1992gp} commonly used for SGP. This proved to be beneficial since GOM varies nodes instead of subtrees~\cite{virgolin2019model,virgolin2018gecco}. For both HH and RHH, syntactical uniqueness of solutions is enforced for up to 100 tries~\cite{koza1992gp}. 
In GP-GOMEA\textsubscript{RT} we additionally avoid sampling trees having a terminal node as root by setting the minimum tree height of the grow method to 1. This is not done for SGP and SGP\textsubscript{b}, because differently from GP-GOMEA\textsubscript{RT} where homologous nodes are varied, subtree root nodes for subtree crossover (SX) and subtree mutation (SM) are chosen uniformly randomly. RS samples new trees using the same initialization method as SGP\textsubscript{b}, i.e., RHH. 

The division operator $\div$ used in the function set is the analytic quotient operator ($ a \div b = a / \sqrt{1 + b^2}$), which was shown to lead to better generalization performance than protected division~\cite{ni2013use}. The logarithm is protected $\log_p(\cdot) = \log( | \cdot | )$ and $\log_p(0)=0$, and so is the square root operator. The terminal set contains the original feature set, and an Ephemeral Random Constant (ERC)~\cite{poli2008field} with values uniformly sampled between the minimum and maximum values of the features in the original training set, i.e.,  $[\min x^{(i)}_j, \max x^{(i)}_j], \forall i,j \in Tr$.

The hyperparameter settings for the SVM, RF and XGB are shown in Table~\ref{tab:ML algorithm-parameters}, and are mostly default~\cite{breiman2001random,CC01a,wright2015ranger} (for XGB, we referred to https://bit.ly/2JCM9x4). NB and LR implementations do not have hyperparameters.

We consider 10 classification and 10 regression benchmark datasets\footnote{The datasets are available at http://goo.gl/9D2z3b} that can be considered traditional, i.e, they have small to moderate dimensionality (number of features). We mostly study this type of dataset because we seek to find small constructed features that can be interpreted. Hence, they can represent a transformation of only a limited number of original features. 
Details on the datasets are reported in Table~\ref{tab:datasets}. Rows with missing values are omitted. Most datasets are taken from the UCI Machine Learning repository\footnote{http://archive.ics.uci.edu/ml/}, with exception for Dow Chemical and Tower, which come from GP literature~\cite{white2013better,albinati2015effect}.

We further consider a very high-dimensional dataset from UCI (https://bit.ly/334KbgW) to assess whether GP can still be useful to construct features in this type of scenario. The dataset in question concerns the classification of cancer type, given RNA-Seq gene expression levels as features. Five cancer class types are present, and class proportions in the data presents some unbalance: the class frequencies are 0.37, 0.18, 0.18, 0.17, 0.10. A total of $20,531$ features are considered, in $801$ examples. Since large computational resources are needed to handle this dataset, we consider only NB as ML algorithm for feature construction upon this data.

\begin{table}
\caption{Salient hyper-parameter settings of SVM, RF, and XGB.}
\label{tab:ML algorithm-parameters}
\small
\centering
\scalebox{0.80}{
\begin{tabular}{lc}
\toprule
\multicolumn{2}{c}{SVM}\\
\midrule
Kernel & RBF \\
Cost & 1 \\
Epsilon & 0.1 \\
Tolerance & 0.001 \\
Gamma & $\frac{1}{k} $ \\
Shrinking & Active \\
\midrule
\multicolumn{2}{c}{RF}\\
\midrule
Number of trees & 100 \\
Bagging sampling & with replacement \\
Classification mtry & $\sqrt{\text{\#features}}$ \\
Regression mtry & $\min ( 1, \frac{\text{\#features}}{3} )$ \\	
Min node size & $1$ classification, $5$ regression \\
Split rule & Gini classification, Variance regression \\
\midrule
\multicolumn{2}{c}{ XGB }\\
\midrule
Number of trees & 100 \\
Booster & gbtree \\
Max depth & 6 \\
Objective & multiclass softmax, MSE regression\\
Learning rate & 0.3\\
\bottomrule
\end{tabular}
}
\end{table}

\begin{table}
\caption{Traditional classification and regression datasets.}
\label{tab:datasets}
\small
\centering
\scalebox{0.75}{
\begin{tabular}{clccc}
\toprule
 & Dataset & \# Features & \# Examples & \# Classes \\
 \midrule
 \parbox[t]{2mm}{\multirow{10}{*}{\rotatebox[origin=c]{90}{Classification}}} 
 & Cylinder Bands & 39 & 277 & 2\\
 & Breast Cancer Wisc. & 29 & 569 & 2\\
 & Ecoli & 7 & 336 & 8\\
 & Ionosphere & 34 & 351 & 2\\
 & Iris & 4 & 150 & 3\\
 & Madelon & 500 & 2600 & 2\\
 & Image Segmentation & 19 & 2310 & 7\\
 & Sonar & 60 & 208 & 2\\
 & Vowel & 9 & 990 & 11\\
 & Yeast & 8 & 1484 & 10\\
\midrule
 \parbox[t]{2mm}{\multirow{10}{*}{\rotatebox[origin=c]{90}{Regression}}} 
 & Airfoil & 6 & 1503 & -- \\
 & Boston Housing & 13 & 506 & --\\
 & Concrete & 9 & 1030 & --\\
 & Dow Chemical & 57 & 1066 & --\\
 & Energy Cooling & 9 & 768 & --\\
 & Energy Heating & 9 & 768 & --\\
 & Tower & 26 & 4999 & --\\
 & Wine Red & 12 & 1599 & --\\
 & Wine White & 12 & 4898 & --\\
 & Yacht & 7 & 308 & --\\
\bottomrule
\end{tabular}
}
\end{table}

\section{Results: performance on traditional datasets}\label{sec:result-rq12}
The results described in this section aim at assessing whether it is possible to construct few and small features that lead to an equal or better performance than the original set, and whether some search algorithms can construct better features than others. 

\subsection{General performance of feature construction}\label{sec:performancefcs}
We begin by observing the dataset-wise aggregated performance of FCS for the different GP algorithms and the different ML algorithms, separately for classification and regression.

\begin{figure*}
\centering

\tabcolsep=0.00mm
\scalebox{1.0}{
\begin{tabular}{
m{0.02\textwidth}
m{0.13425\textwidth}m{0.12\textwidth}|m{0.12\textwidth}
m{0.12\textwidth}|m{0.12\textwidth}m{0.12\textwidth}|m{0.12\textwidth}m{0.12\textwidth}
}
& \multicolumn{2}{c}{ NB } & \multicolumn{2}{c}{ SVM } & \multicolumn{2}{c}{RF} & \multicolumn{2}{c}{XGB}\\[-1.0mm] 
& \begin{center} Training \end{center} & \begin{center} Test \end{center} & \begin{center} Training \end{center} & \begin{center} Test \end{center} & \begin{center} Training \end{center} & \begin{center} Test \end{center} & \begin{center} Training \end{center} & \begin{center} Test \end{center}\\[-4mm] 
 
\begin{sideways} $h=2$ \end{sideways} & 
\includegraphics[width=\linewidth]{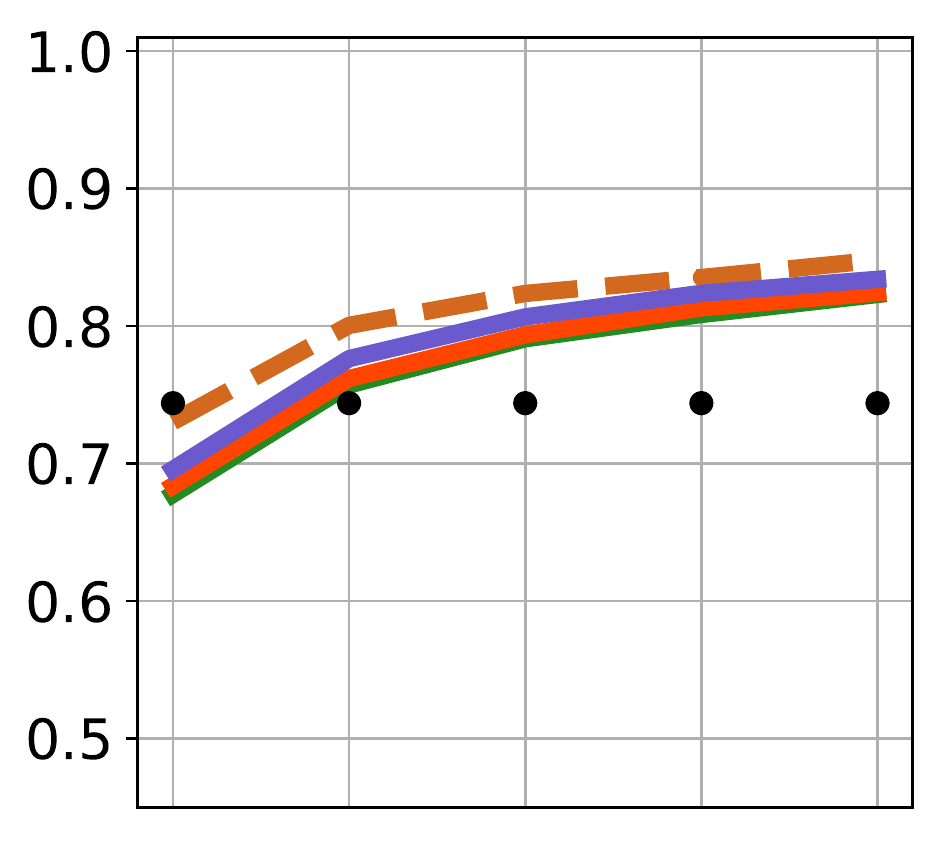} &
\includegraphics[width=\linewidth]{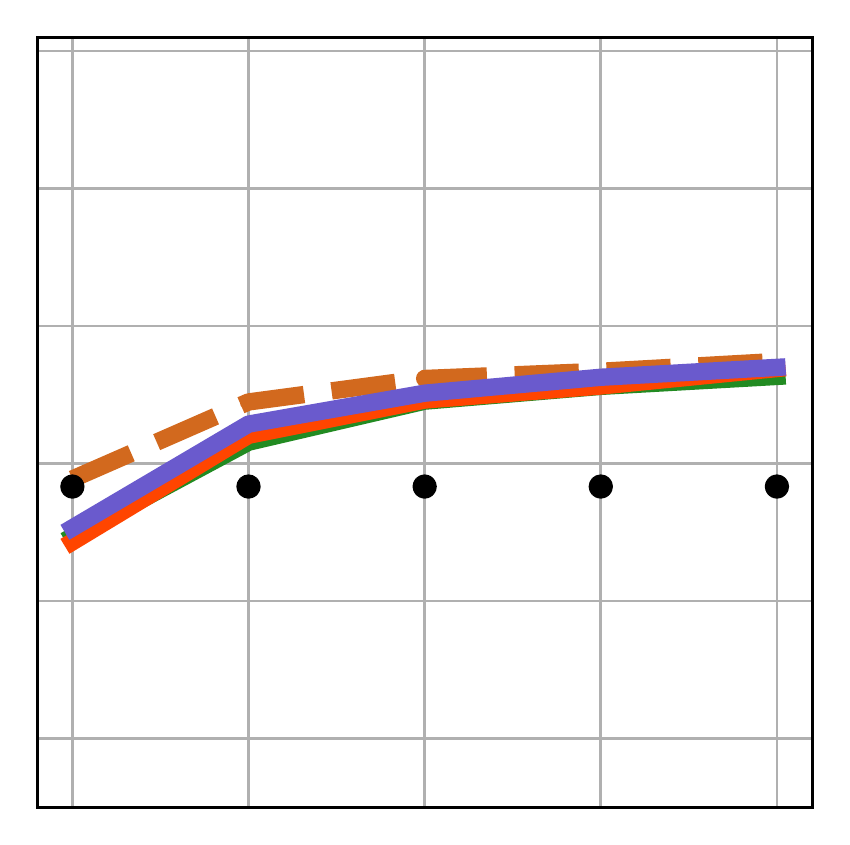} &
\includegraphics[width=\linewidth]{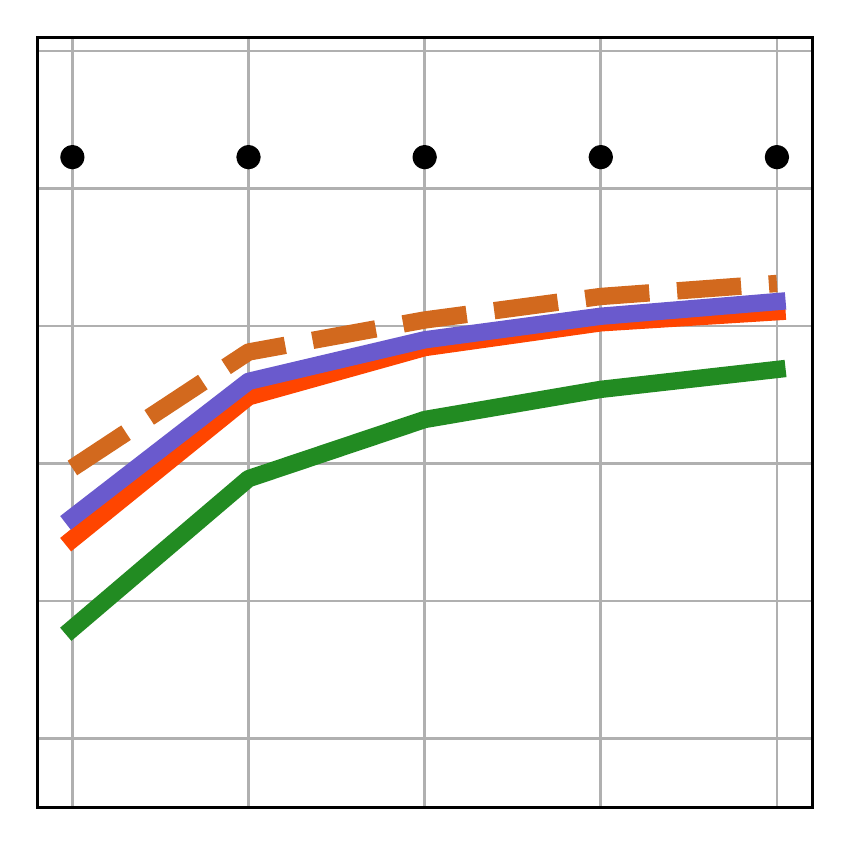} &
\includegraphics[width=\linewidth]{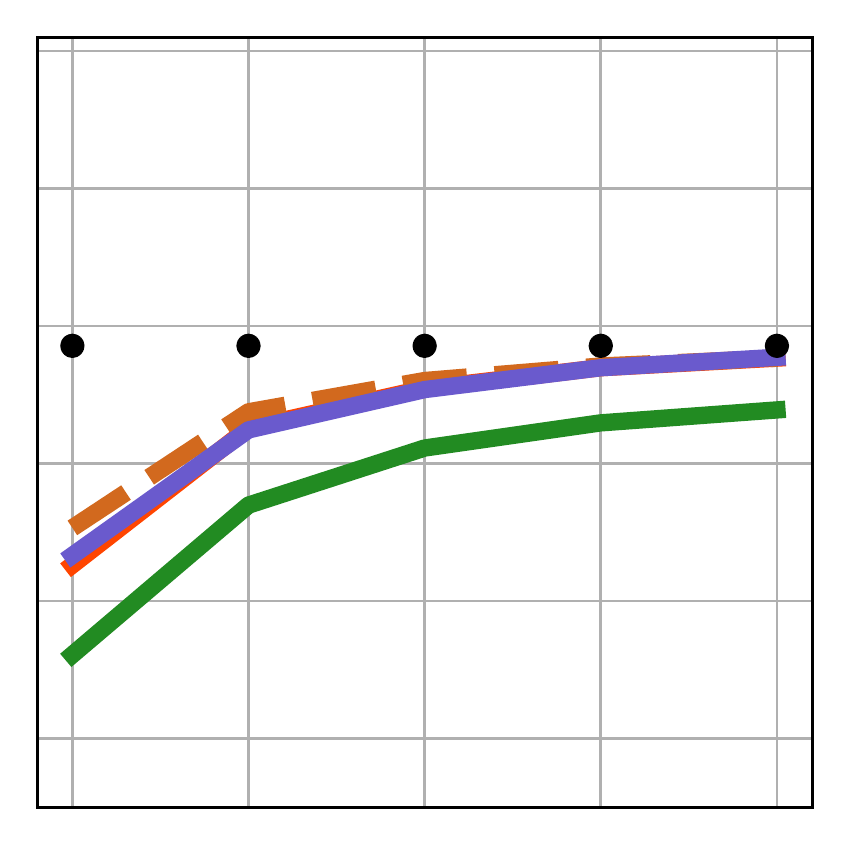} &
\includegraphics[width=\linewidth]{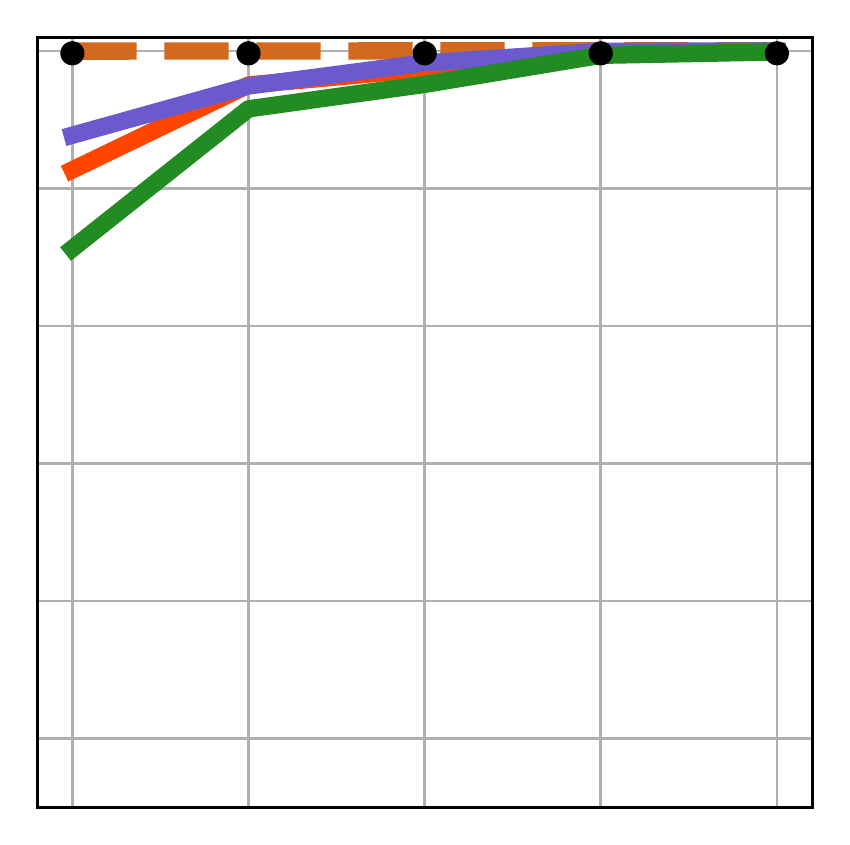} &
\includegraphics[width=\linewidth]{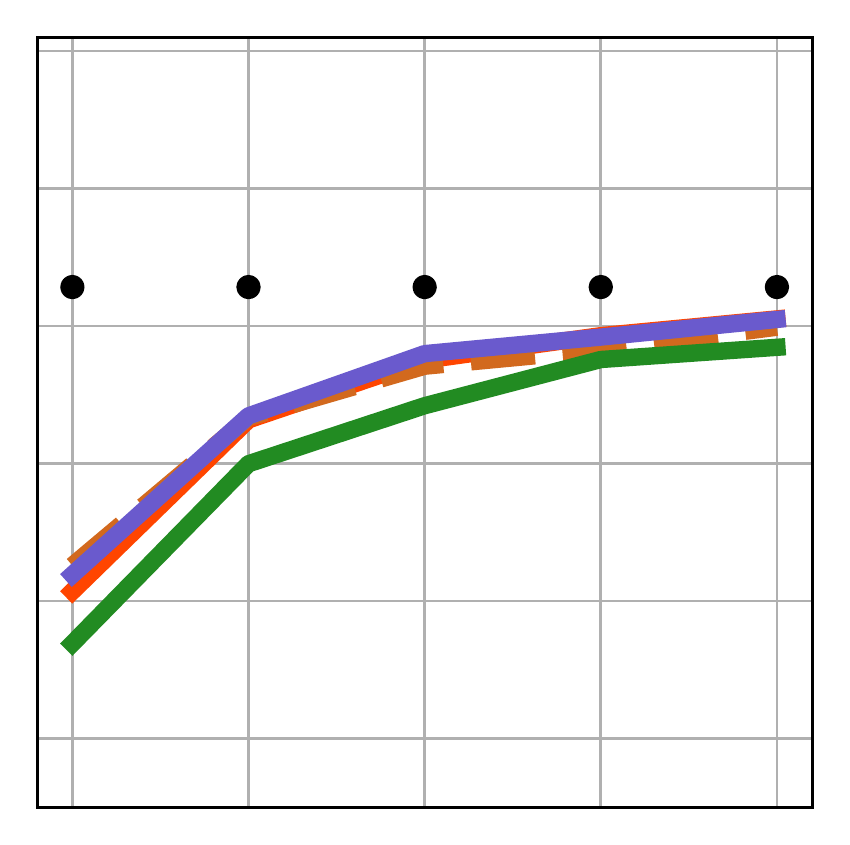} &
\includegraphics[width=\linewidth]{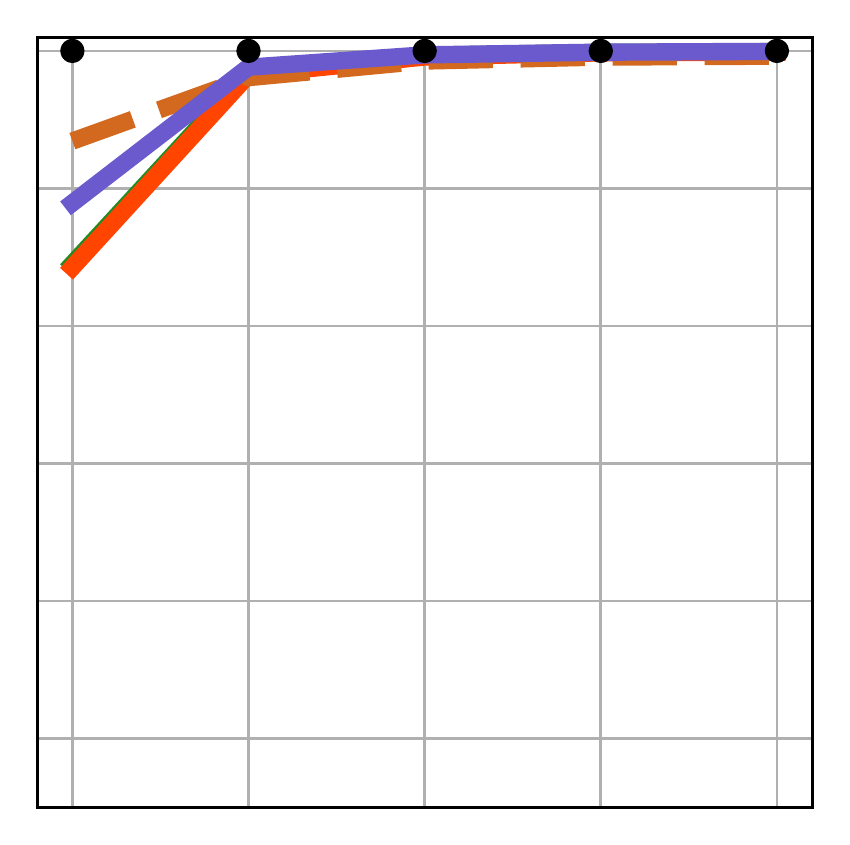} &
\includegraphics[width=\linewidth]{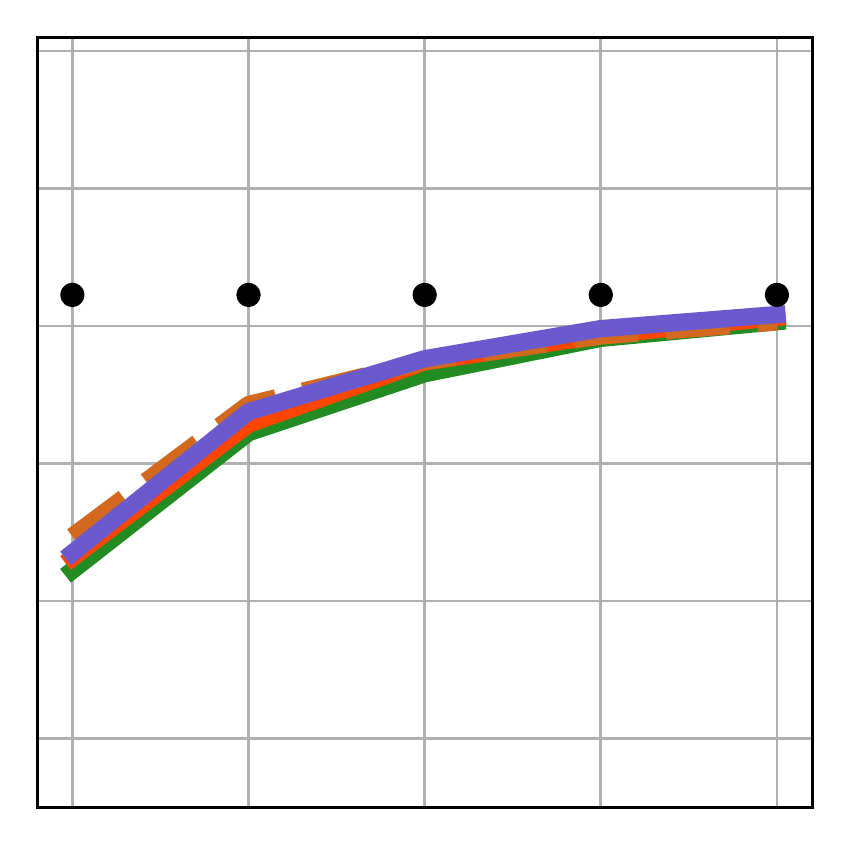}
\\[-1.0mm]

\begin{sideways}  $h=4$ \end{sideways} & 
\includegraphics[width=\linewidth]{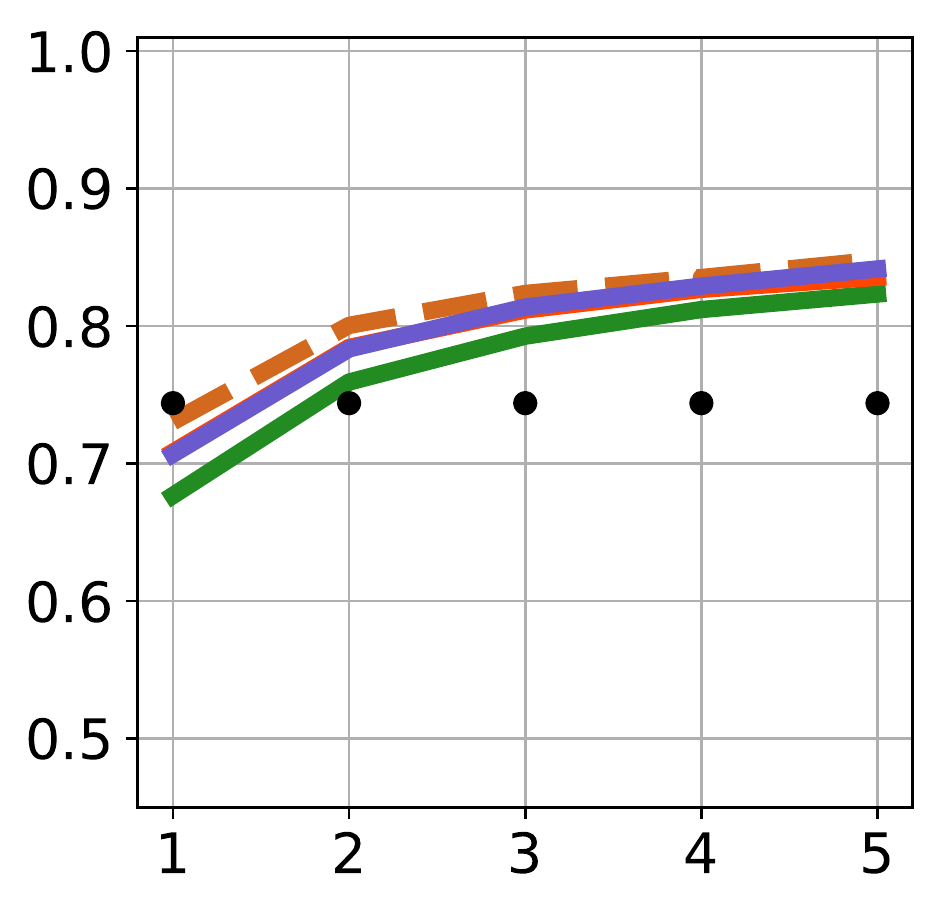} &
\includegraphics[width=\linewidth]{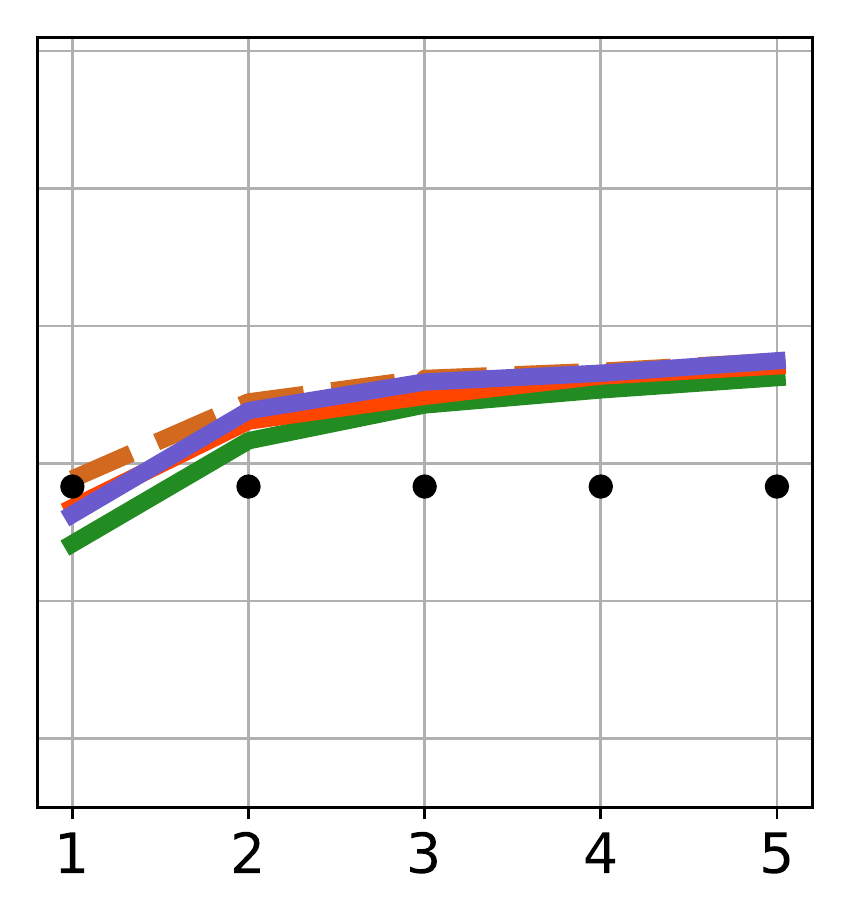} &
\includegraphics[width=\linewidth]{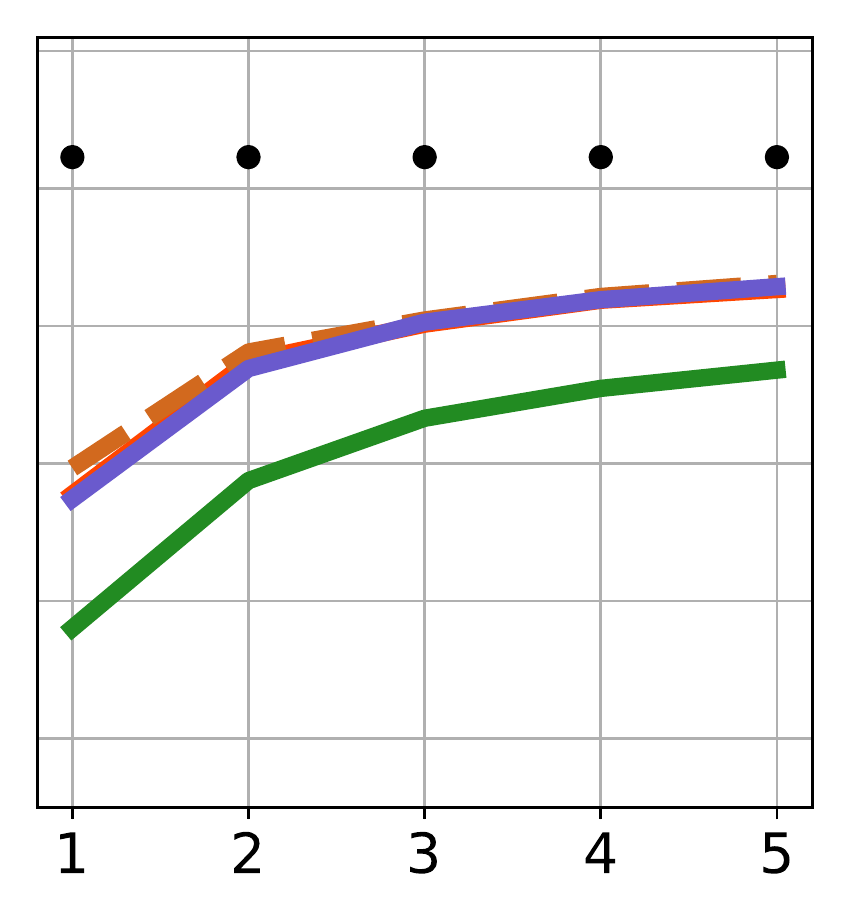} &
\includegraphics[width=\linewidth]{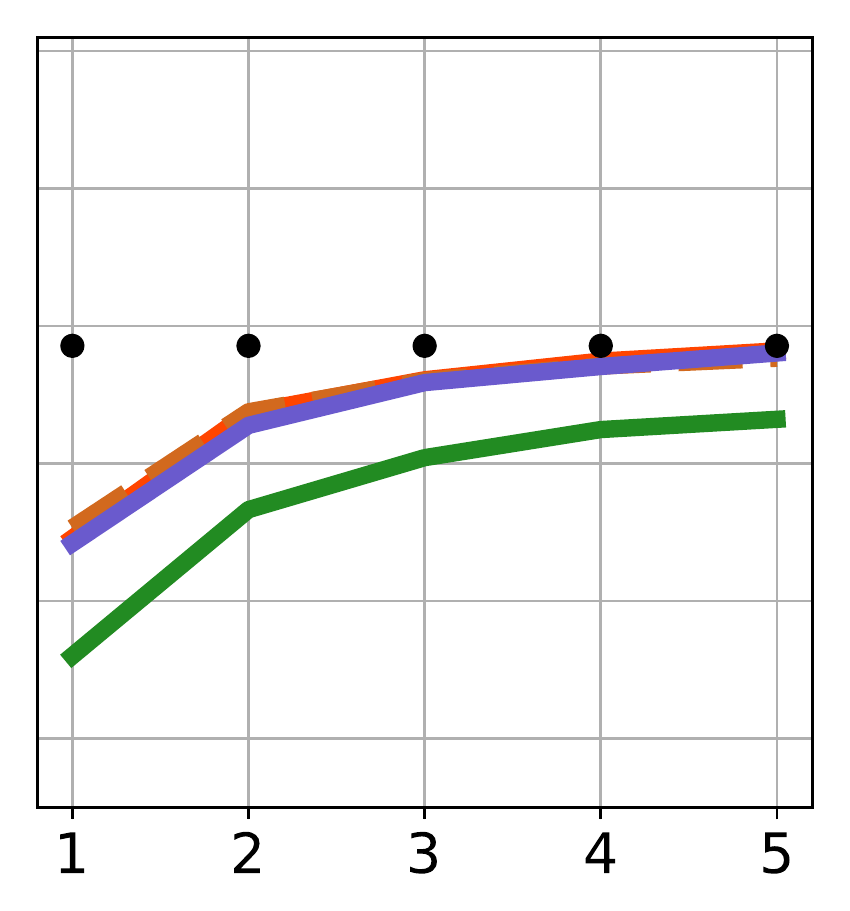} &
\includegraphics[width=\linewidth]{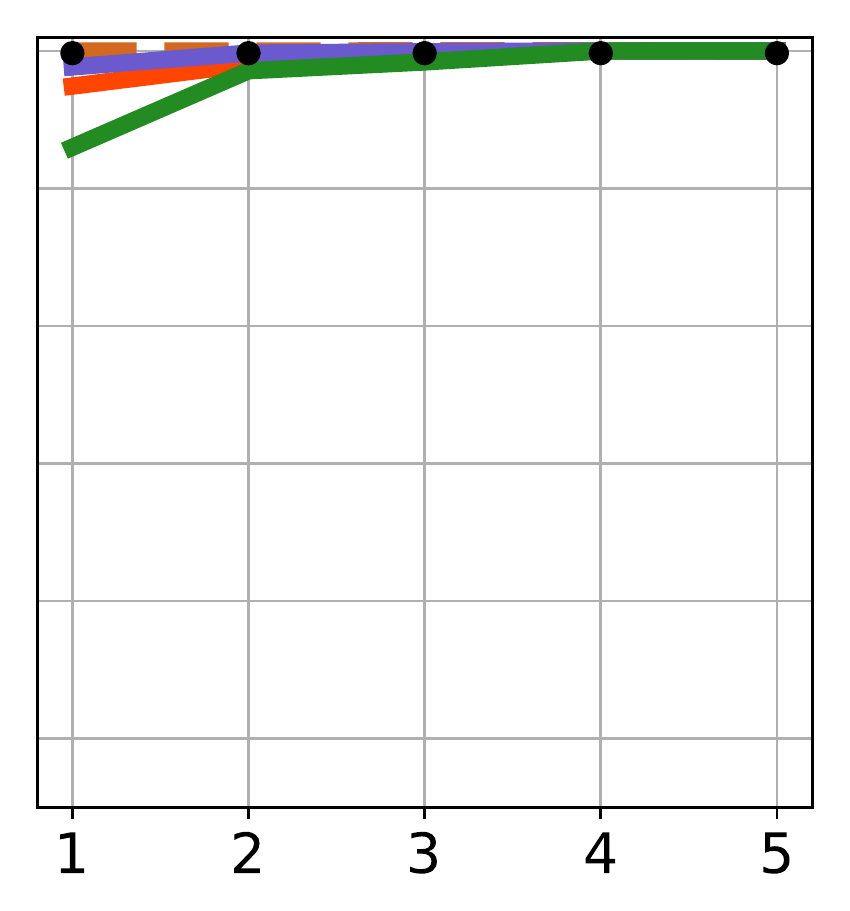} &
\includegraphics[width=\linewidth]{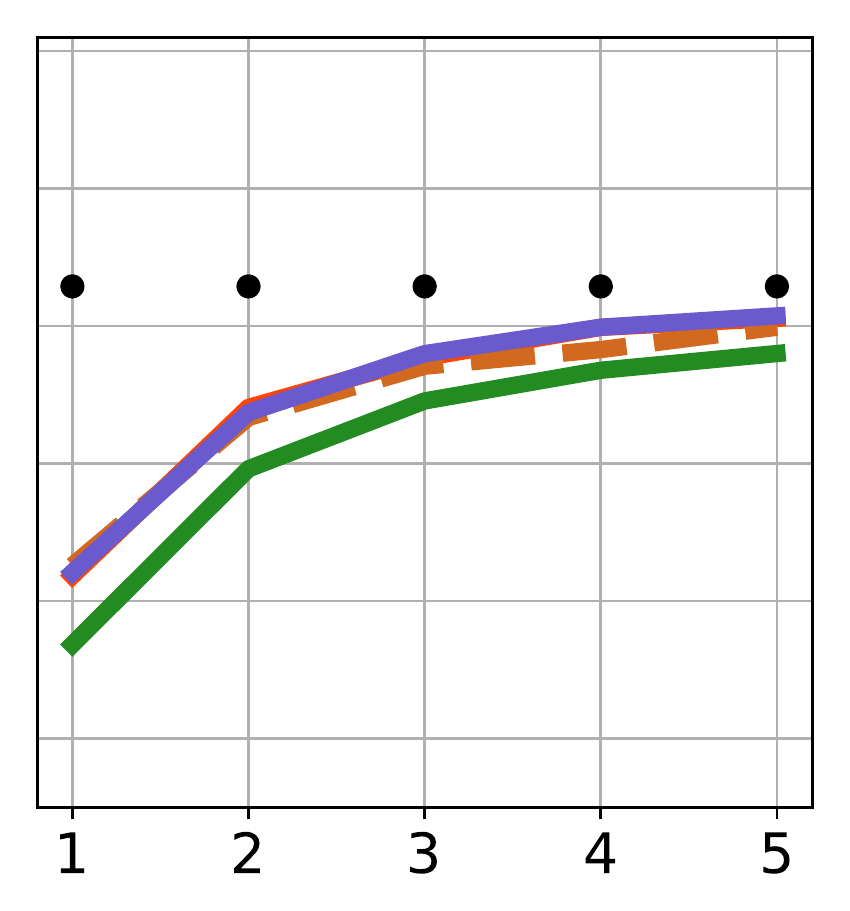} &
\includegraphics[width=\linewidth]{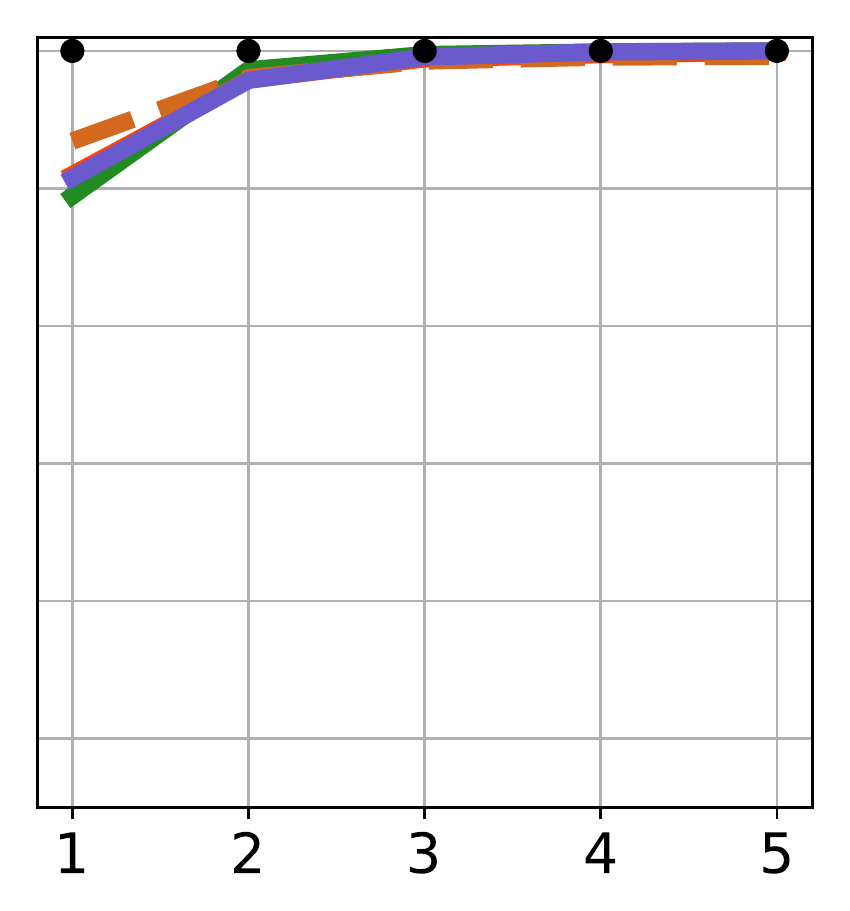} &
\includegraphics[width=\linewidth]{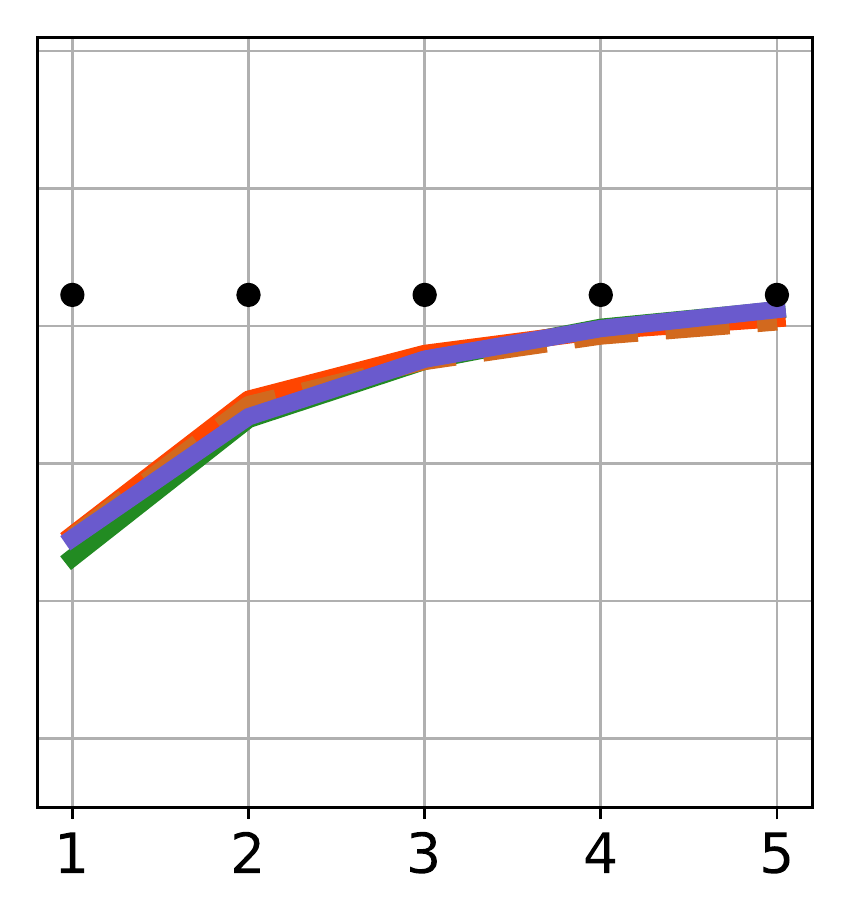}
\\

\end{tabular}
}
\includegraphics[height=0.3cm]{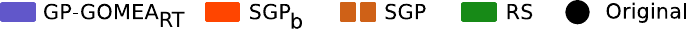}

\caption{Aggregated results on the classification datasets. Horizontal axis: Number of features. Vertical axis: Average of median $F1$ score obtained on 30 runs for each dataset.}\label{fig:means_median_cls}

\end{figure*}

\begin{figure*}
\centering

\tabcolsep=0.00mm
\scalebox{1.0}{
\begin{tabular}{
m{0.02\textwidth}
m{0.13425\textwidth}m{0.12\textwidth}|m{0.12\textwidth}
m{0.12\textwidth}|m{0.12\textwidth}m{0.12\textwidth}|m{0.12\textwidth}m{0.12\textwidth}
}
& \multicolumn{2}{c}{ LR } & \multicolumn{2}{c}{ SVM } & \multicolumn{2}{c}{RF} & \multicolumn{2}{c}{XGB }\\[-1.0mm] 

& \begin{center} Training \end{center} & \begin{center} Test \end{center} & \begin{center} Training \end{center} & \begin{center} Test \end{center} & \begin{center} Training \end{center} & \begin{center} Test \end{center} & \begin{center} Training \end{center} & \begin{center} Test \end{center}\\[-4mm]

\begin{sideways} $h=2$ \end{sideways} & 
\raisebox{0.0mm}{\includegraphics[width=\linewidth]{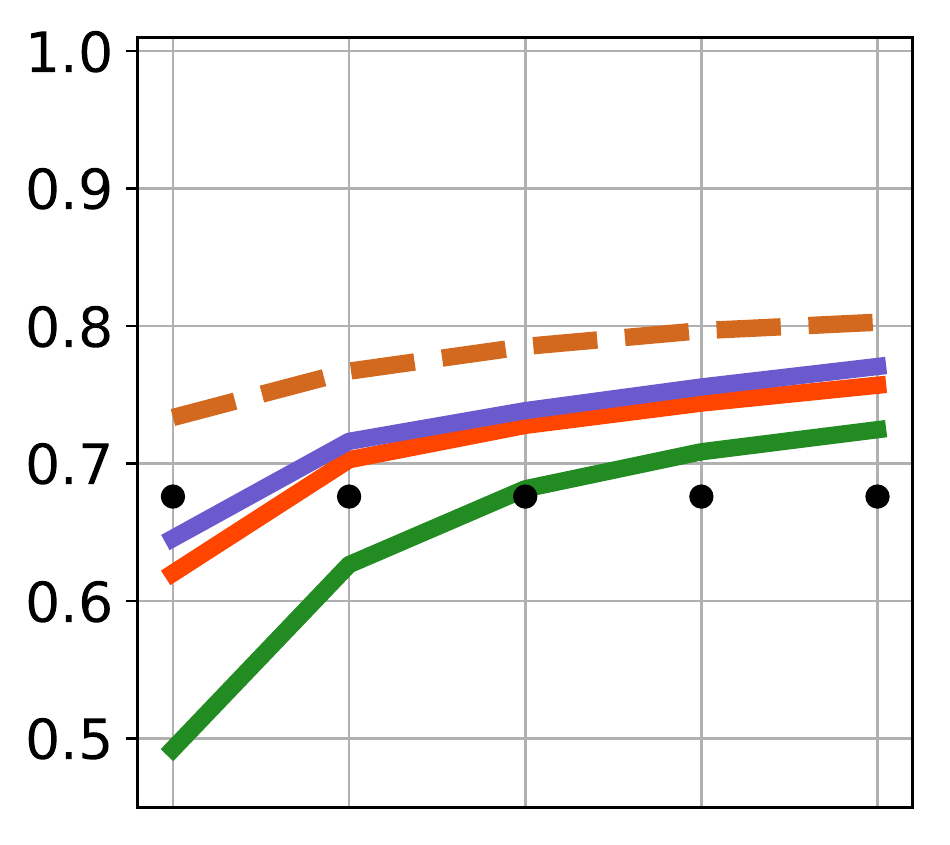}} &
\includegraphics[width=\linewidth]{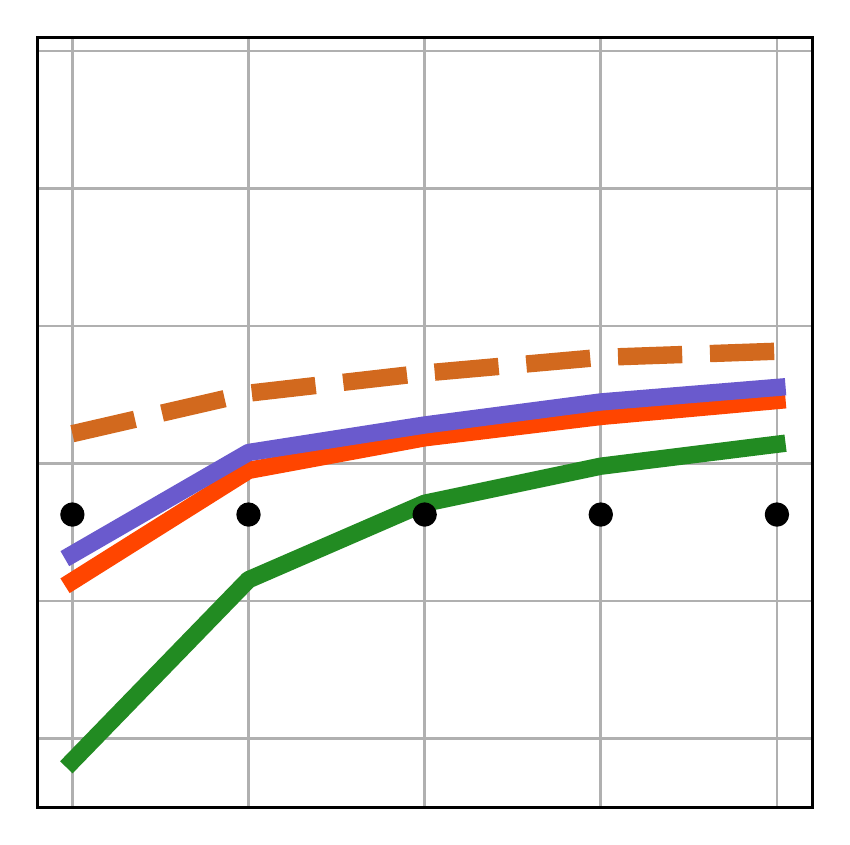}&
\includegraphics[width=\linewidth]{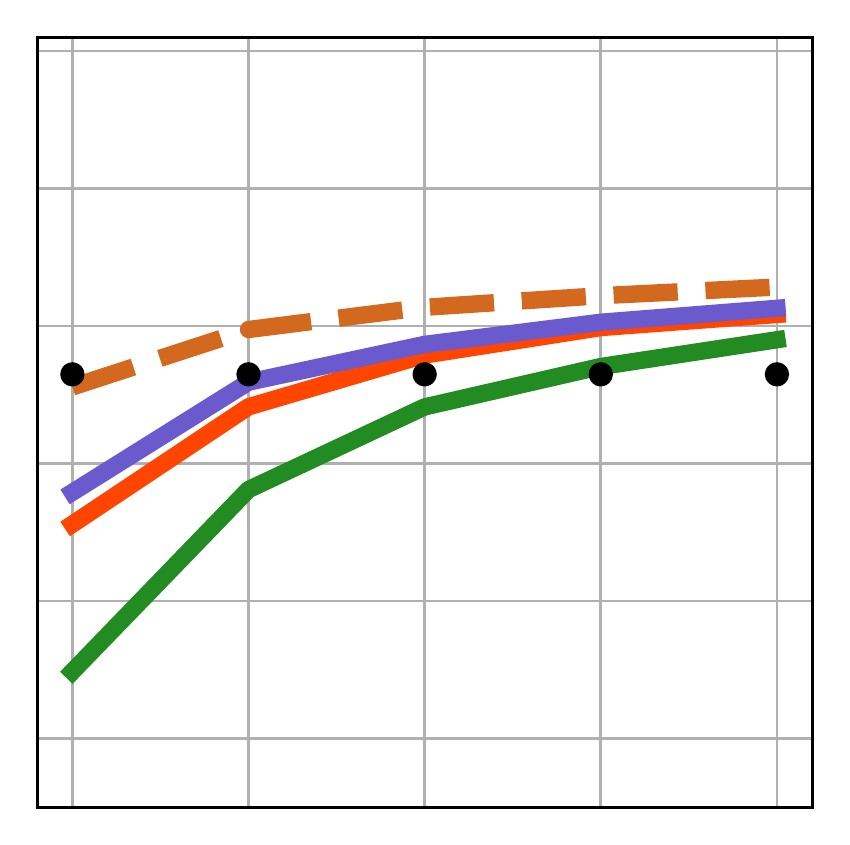}&
\includegraphics[width=\linewidth]{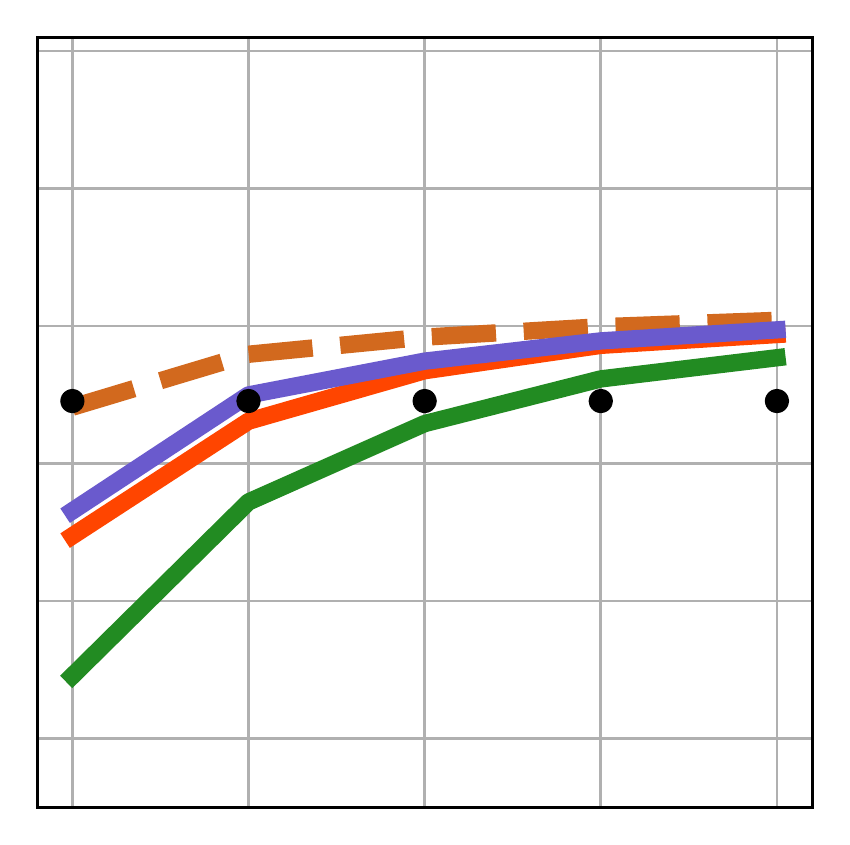}&
\includegraphics[width=\linewidth]{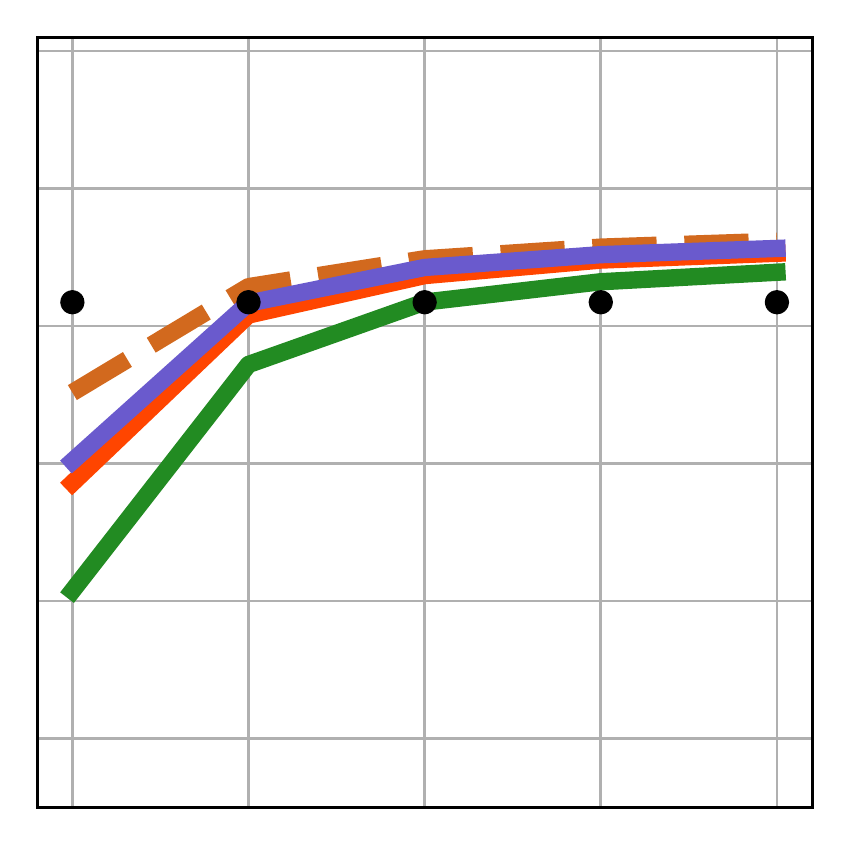} &
\includegraphics[width=\linewidth]{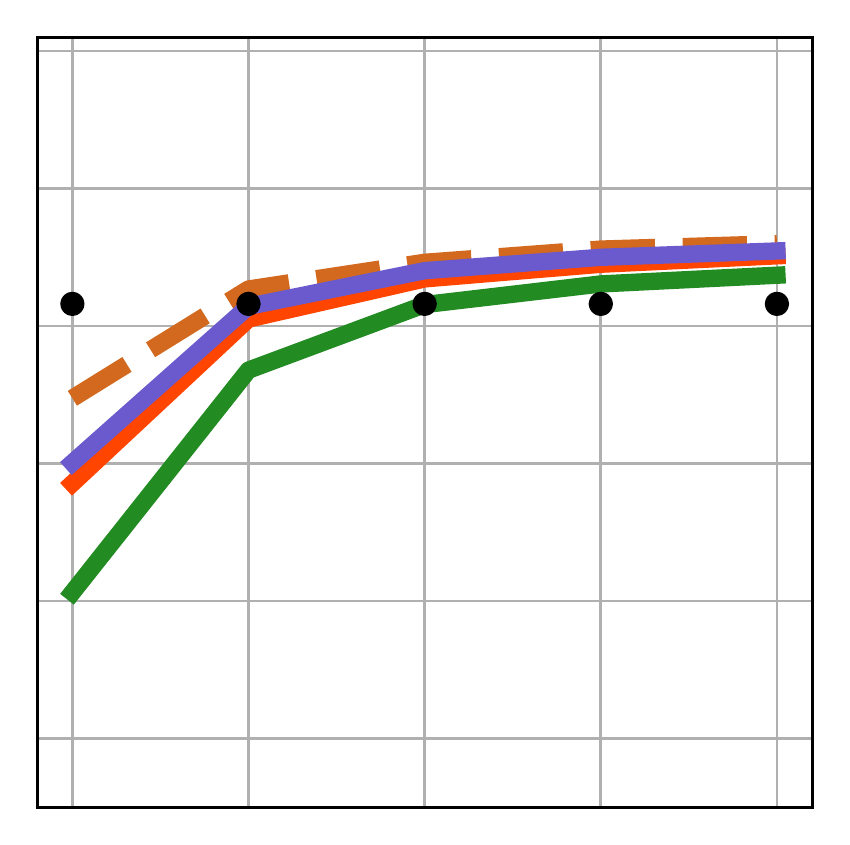} &
\includegraphics[width=\linewidth]{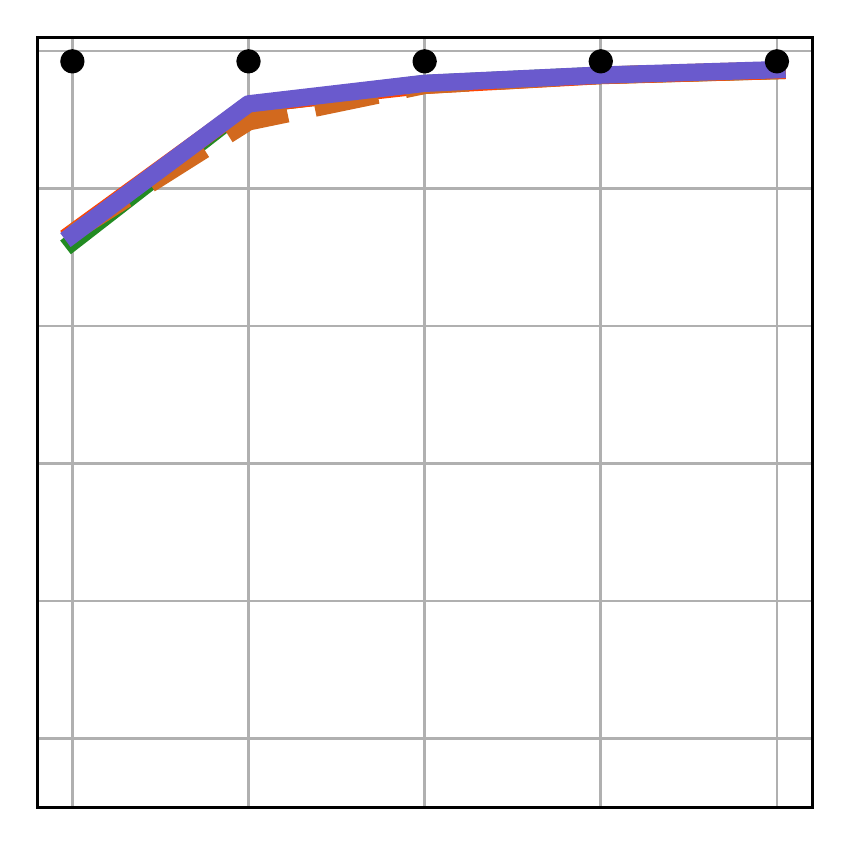} &
\includegraphics[width=\linewidth]{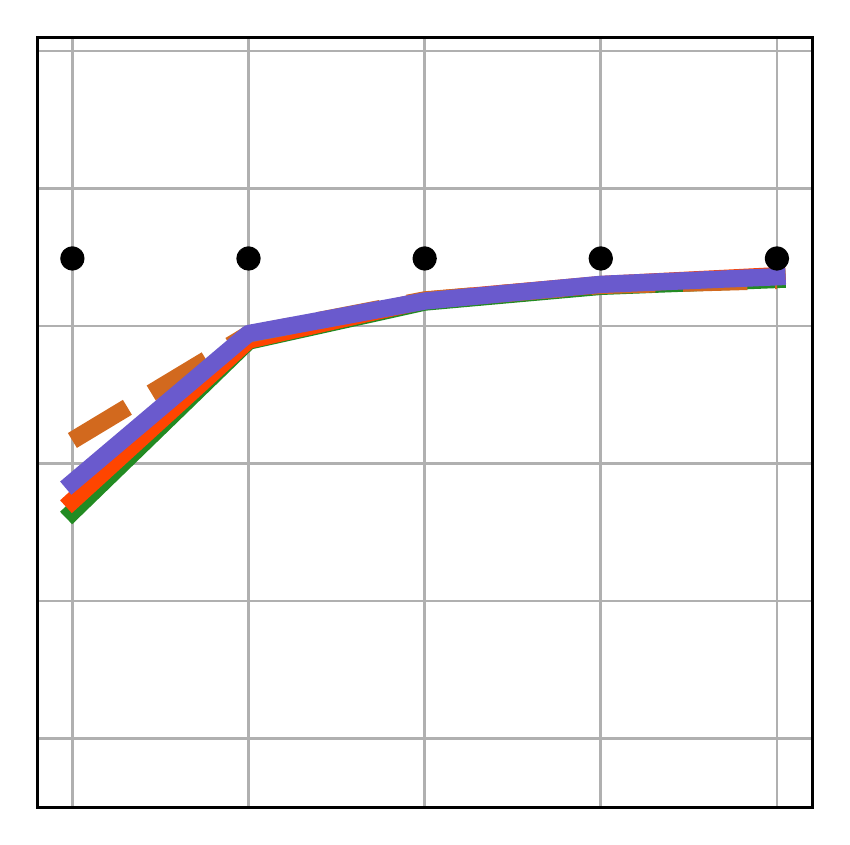}
\\[-1.0mm]

\begin{sideways}  $h=4$ \end{sideways} & 
\includegraphics[width=\linewidth]{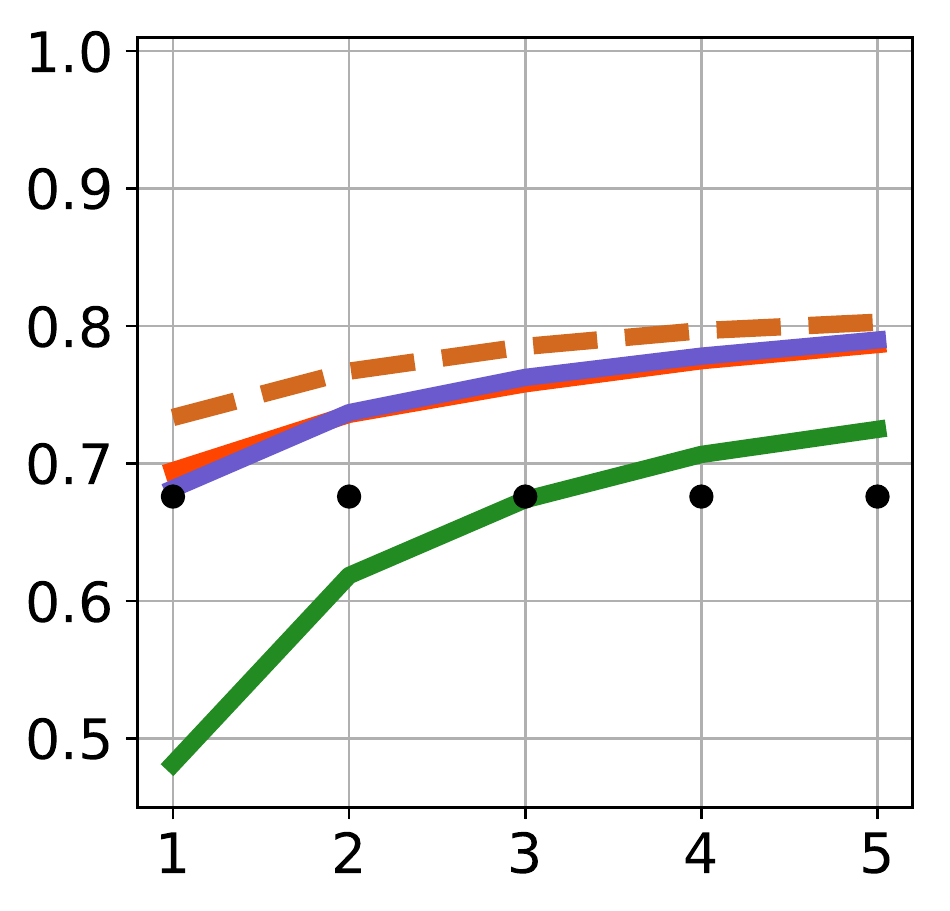}&
\includegraphics[width=\linewidth]{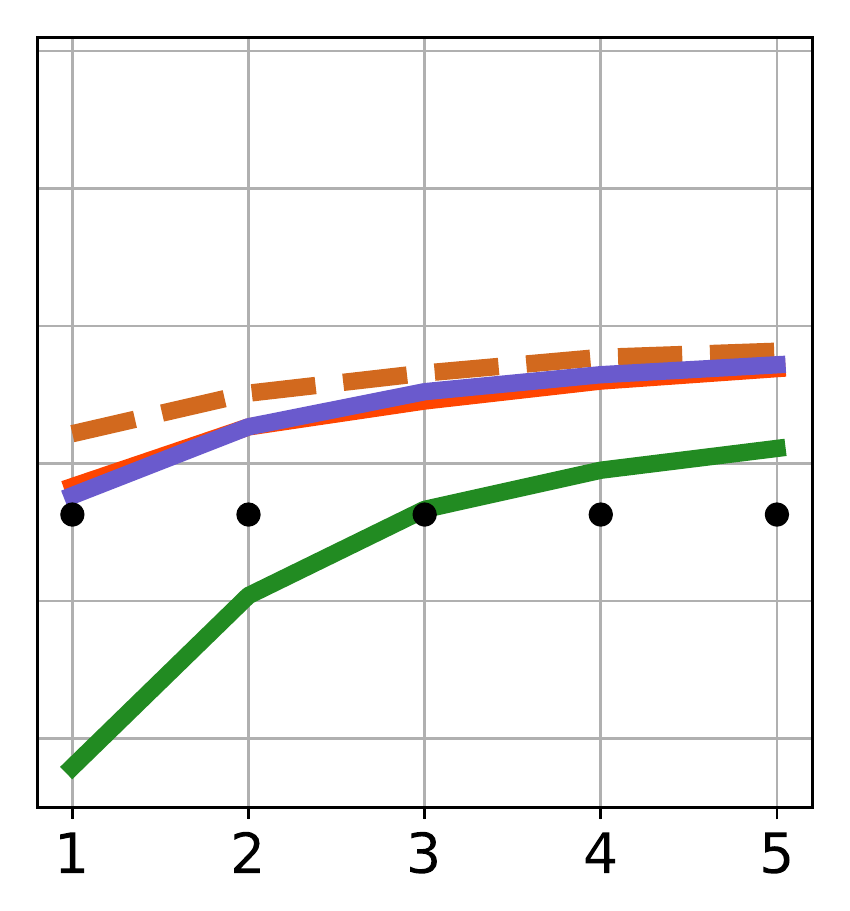}&
\includegraphics[width=\linewidth]{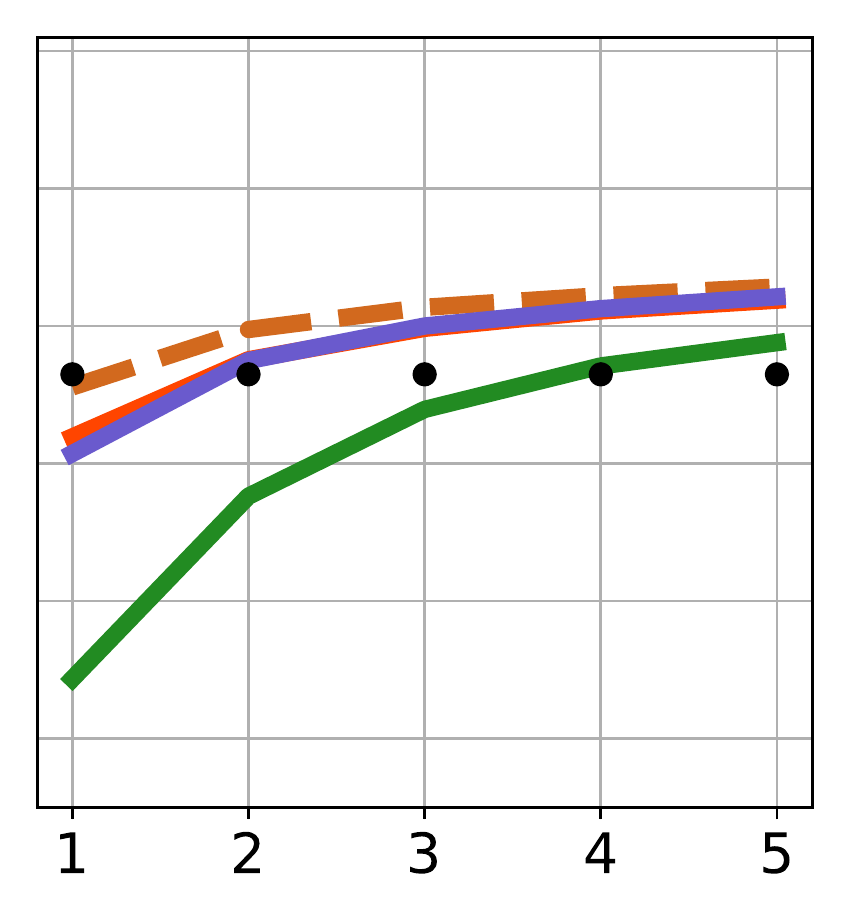}&
\includegraphics[width=\linewidth]{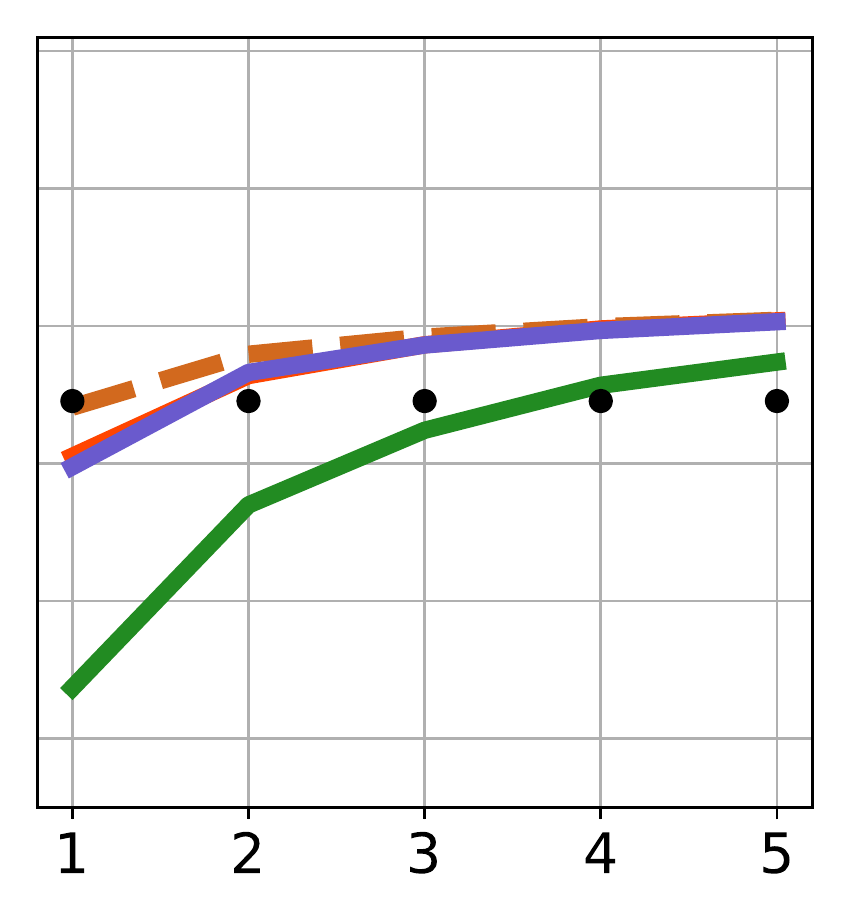}&
\includegraphics[width=\linewidth]{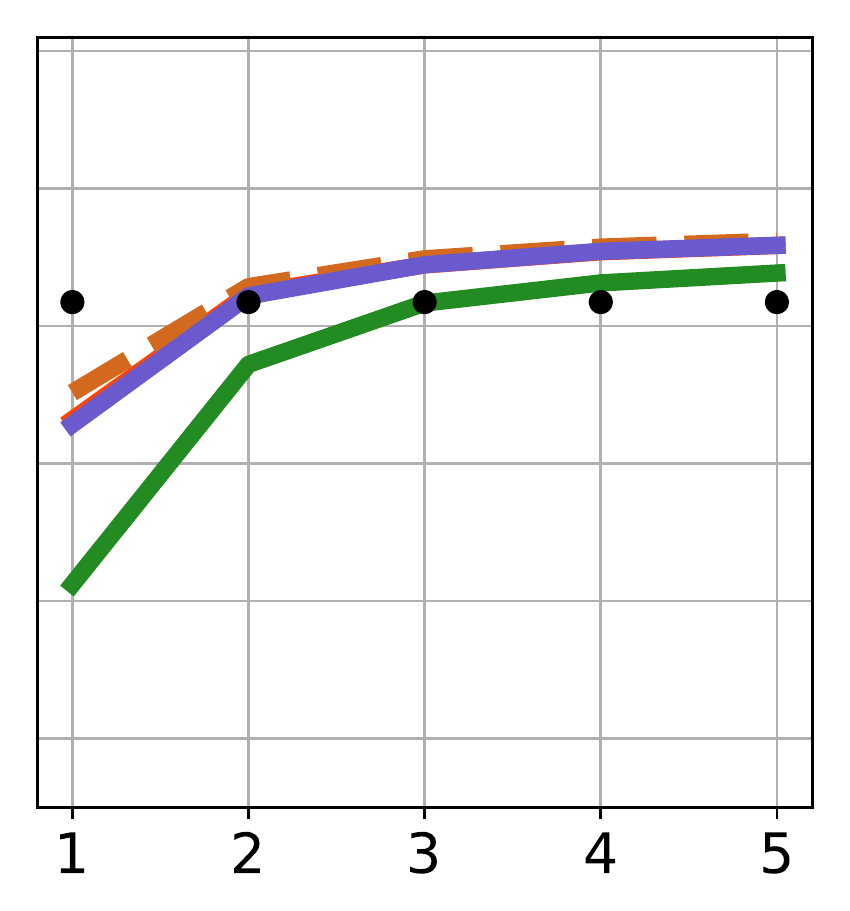}&
\includegraphics[width=\linewidth]{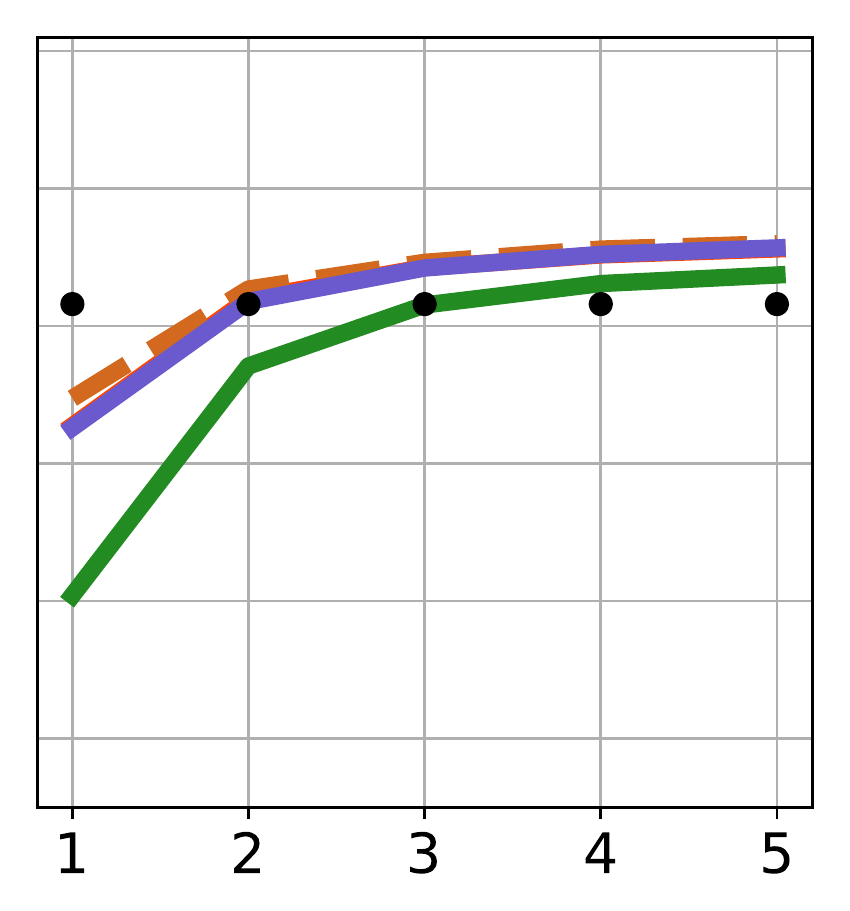} &
\includegraphics[width=\linewidth]{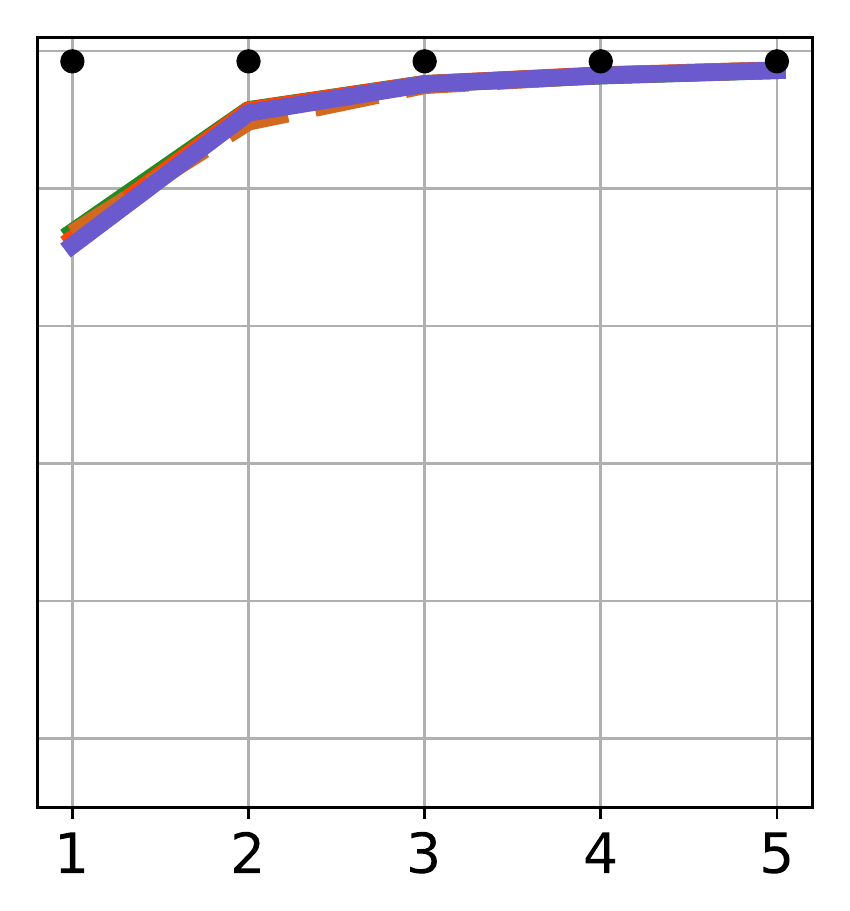} &
\includegraphics[width=\linewidth]{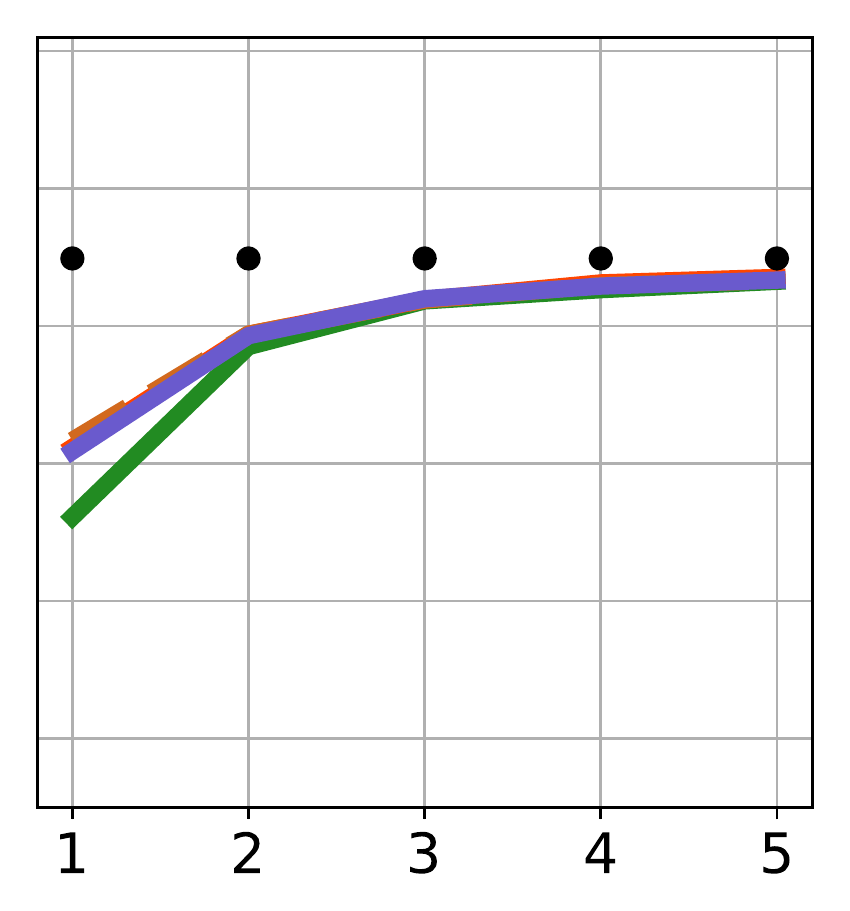}
\\

\end{tabular}
}
\includegraphics[height=0.3cm]{legend}

\caption{Aggregated results on the regression datasets. Horizontal axis: Number of features. Vertical axis: Average of median $R^2$ score obtained on 30 runs for each dataset.}\label{fig:means_median_regr}

\end{figure*}

\subsubsection{Classification}\label{sec:result-rq12-class} Figure~\ref{fig:means_median_cls} shows dataset-wise aggregated results obtained for NB, SVM, RF, and XGB, for the 10 traditional classification tasks. 
Each data point is the mean among the dataset-specific medians of macro F1 from the 30 runs.

In general, the use of only one constructed feature does not perform as good as the use of the original feature set. Constructing more features improves the performance, but with diminishing returns.

Specifically for NB, the use of two constructed features is already preferable to the use of the original feature set. This is likely due to the fact that NB assumes complete independence between the provided features, and this can be implicitly tackled by FCS. SGP (unbounded) is the best performing algorithm as it can evolve arbitrarily complex features, however, the magnitude of improvement of the macro F1 score with respect to GP-GOMEA\textsubscript{RT} and SGP\textsubscript{b} is limited. For $h=4$ and $K=5$, GP-GOMEA\textsubscript{RT} reaches the performance of SGP. GP-GOMEA\textsubscript{RT} is typically slightly better than SGP\textsubscript{b}, and RS has worse performance. Training and test F1 scores do not differ much for any feature construction algorithm, meaning that overfitting is not an issue for NB. Rather, compared to the other ML algorithms, NB underfits.

The performance of FCS for SVM has an almost identical pattern to the one observed for NB, except for the fact that the performance is found to be consistently better. However, for SVM it is preferable to use the original feature set rather than few constructed features. This is evident in terms of training performance, but less at test time. In fact, using only 5 constructed features leads to similar test performance compared to using the original set. The GP algorithms compare to each other similarly to when using NB. Compared to NB, it can be seen that SVM exhibits larger gaps between training and test results, suggesting that some overfitting takes place, especially when the original feature set is used.

The way performance improves for RF by constructing features is similar to the one observed for NB and SVM. However, for RF the differences between the search algorithms is particularly small: notice that using RS leads to close performance to the ones obtained by using the other GP algorithms, compared to the SVM case. Moreover, virtually no difference can be seen between GP-GOMEA\textsubscript{RT} and SGP\textsubscript{b}. This suggests that RF already works well with less refined features. Now, the features constructed by SGP are no longer the best performing at test time. This is likely because SGP evolves larger, more complex features than the other algorithms (see Sec.~\ref{sec:feature-size}), making RF overfit. In fact, RF exhibits the largest difference between training and test results compared to NB and SVM, for any feature construction algorithm and $h$ limit. Still, the test results of RF are slightly better than the ones of SVM and markedly better than the ones of NB, meaning that the latter two are underfitting.

The training and test performance obtained when using XGB is similar to the one obtained when using RF, but the differences the between different search algorithms are even less marked than for RF. Some differences can be seen for $K=1$ on the training set (SGP better than GP-GOMEA\textsubscript{RT}, and GP-GOMEA\textsubscript{RT} better than SGPb and RS), but this difference is much less marked on the test set. When more features are constructed, essentially all search algorithms deliver the same performance. XGB seems to be able to construct non-linear relationships even better than RF. As to potential overfitting, the trend of differences between training and test performance that can be observed for XGB mirrors the one visible for RF.

As to maximum tree height, allowing the constructed features to be bigger ($h=4$ vs $h=2$) moderately improves the performance. Interestingly, GP-GOMEA\textsubscript{RT} with $h=4$ reaches competitive performance with SGP on all ML algorithms, despite the latter having no strict limitation on feature size. 

\subsubsection{Regression}\label{sec:result-rq12-regr} Results on the regression tasks are shown in Figure~\ref{fig:means_median_regr}, dataset-wise aggregated for LR, SVM, RF, and XGB. We report the results in terms of coefficient of determination, i.e., $R^2(y, \bar{y}) = 1 - MSE(y, \bar{y})/var(y)$. For the four ML algorithms, results overall follow the same pattern. SGP is typically better, especially for LR and SVM, although constructing more features reduces the performance gap with the other GP algorithms. GP-GOMEA\textsubscript{RT} is slightly, yet consistently, the best performing within the maximum tree height limitation of 2, while SGP\textsubscript{b} is visibly preferable only when a single feature is constructed for LR and SVM, for $h=4$. Differently from the classification case, two features are typically enough to reach the performance of the original feature set for all ML algorithms except for XGB. Moreover, for LR, SVM, and RF, the performance between training and test is similar, meaning no considerable overfitting is taking place, no matter the feature construction algorithm used nor the limit of $h$. This however is not the case for XGB, where a large performance gap is encountered. Still, the test performance obtained when using XGB is ultimately slightly better than the obtained for RF.

As for classification, allowing for larger trees results in better performance overall, and reduces the gap between SGP and the other GP algorithms. With XGB, all search algorithms perform similarly.

\subsubsection{Feature size}\label{sec:feature-size} 
\begin{figure}
\centering
\tabcolsep=0.02mm
\begin{tabular}{lc}
\begin{sideways}\begin{centering}\small \hspace{2cm} Feature size\end{centering}\end{sideways}
& \includegraphics[width=0.65\linewidth]{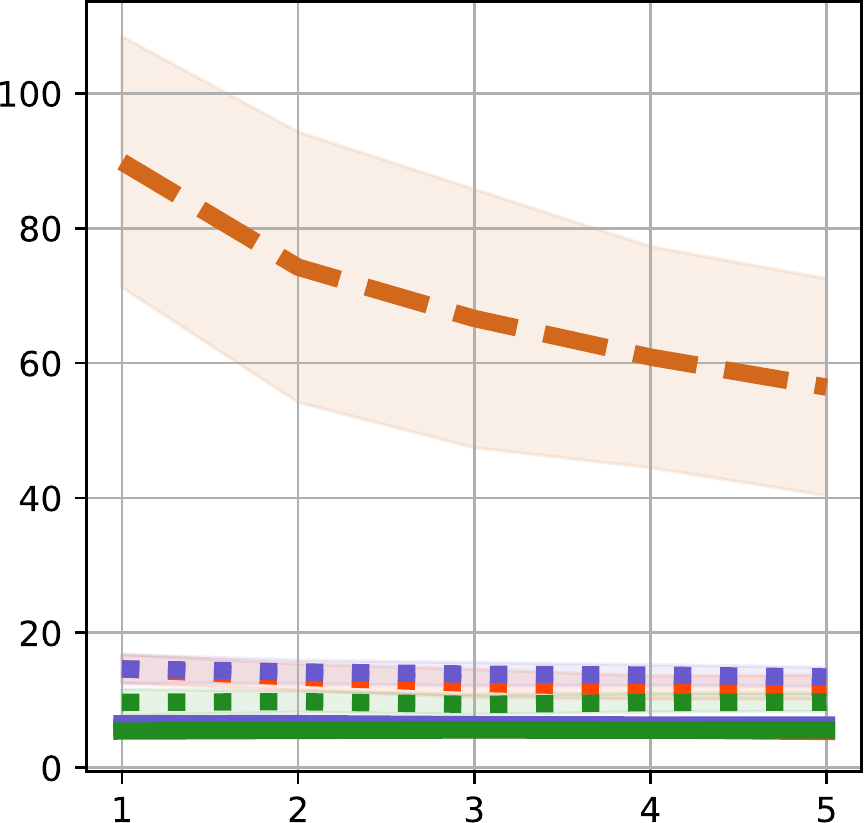}
\\
& \includegraphics[width=0.65\linewidth]{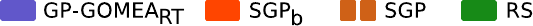}
\end{tabular}
\caption{Aggregated feature size for $k = 1,\dots,5$. Solid (dotted) lines represent solution size for maximum tree height $h=2$ ($h=4$). Shaded areas represent standard deviation. SGP is free to grow solutions up to $h=17$.}\label{fig:size-all}
\end{figure}

Figure~\ref{fig:size-all} shows the aggregated feature size for the different GP algorithms and RS. The aggregated solution size is computed by taking the median solution size per run, then averaging over datasets, and finally averaging over ML algorithms (classification and regression are considered together). The picture shows how, overall, the known SGP tendency to bloat differs compared to the algorithms working with a strict tree height limitation. SGP features are so large that it is nearly impossible to interpret them (see Sec.~\ref{sec:interpretability}).

RS finds the smallest features for both height limits $h=2$ and $h=4$. Considering that GP-GOMEA\textsubscript{RT} and SGP\textsubscript{b} generate trees within the same height bounds of RS, we conclude that it is the variation operators that allow finding larger trees with improved fitness within the height limit. GP-GOMEA\textsubscript{RT} seems to construct slightly, yet consistently, larger trees than SGP\textsubscript{b}.

For SGP, it can be seen that subsequently constructed features are smaller (this is barely visible for GP-GOMEA\textsubscript{RT} and SGP\textsubscript{b} as well). This is interesting because we do not use any mechanism to promote smaller trees. 
This result is likely linked to the diminishing returns in performance observed in Figure~\ref{fig:means_median_cls} and~\ref{fig:means_median_regr}: constructing new complex and informative features becomes harder with the number of FCS iterations.

\subsection{Statistical significance: comparing GP algorithms}
The aggregated results of Section~\ref{sec:performancefcs} show moderate  differences between GP-GOMEA\textsubscript{RT} and SGP\textsubscript{b}. These are arguably the most interesting algorithms to compare in-depth, as they are able to construct small features that lead to good performance (RS typically constructs less informative features, while SGP constructs very large ones).

We perform statistical significance tests to compare GP-GOMEA\textsubscript{RT} and SGP\textsubscript{b}. We consider their median performance on the test set $Te$, obtained by the FCS, and also compare it with the use of the original feature set, for each ML algorithm and each dataset. In our case, the \emph{treatments} of our significance tests are the two search algorithms (i.e., GP-GOMEA\textsubscript{RT} and SGP\textsubscript{b}) and the original feature set, while the \emph{subjects} are the configurations given by pairing ML algorithms and datasets~\cite{demvsar2006statistical}.

We first perform a Friedman test to assess whether differences exists among the use of different treatments (GP algorithms and original feature set) upon multiple subjects (ML algorithm-dataset combinations). As post-hoc analysis, we use the pairwise Wilcoxon signed rank tests, paired by subject (ML algorithm-dataset combination), to see how the treatments compare to each other~\cite{demvsar2006statistical}. We adopt the Holm correction method to prevent reporting false positive results that might have happened due to pure chance~\cite{holm1979simple}.

We consider both $h=2$ and $h=4$, and focus on $K=2$, since consideration of only two constructed features makes interpretation easier, and allows human visualization (see Sec.~\ref{sec:interpretability}). 

\subsubsection{Classification}
For both $h=2,4$, the Friedman test strongly indicates differences between GP-GOMEA\textsubscript{RT}, SGP\textsubscript{b}, and the use of the original feature set ($p\text{-value} \ll 0.05$).

Figure~\ref{fig:signiftests} (top) shows the Holm-corrected $p$-values obtained by the pairwise Wilcoxon tests for classification, where the alternative hypothesis is that the row allows for larger macro F1 scores than the column.
No significant differences between GP-GOMEA\textsubscript{RT} and SGP\textsubscript{b} are found for both $h=2,4$. Both the GP algorithms can deliver constructed features that are competitive with the use of the original feature set. The original feature set is not significantly better than using feature construction. Moreover, for GP-GOMEA\textsubscript{RT} and $h=4$, the hypothesis that feature construction is \emph{not} better than the original feature set can be rejected with a corrected $p$-value below 0.1. The latter result appears to be in contrast with the results from Fig.~\ref{fig:means_median_cls} for SVM, RF and XGB, where it can be seen that the construction of only two features does, on average, lead to slightly worse test results than using the original feature set. Nonetheless, the opposite is true for NB, and with rather large magnitude. A more in-depth analysis on this is provided in Sec.~\ref{sec:statanal-per-mla}.

\subsubsection{Regression}
\begin{figure}
\centering
\tabcolsep=0.02mm
\scalebox{0.9}{
\begin{tabular}{
m{0.05\linewidth}m{0.40\linewidth}m{0.40\linewidth}
}
\begin{sideways}Classification\end{sideways}
&
\includegraphics[width=\linewidth]{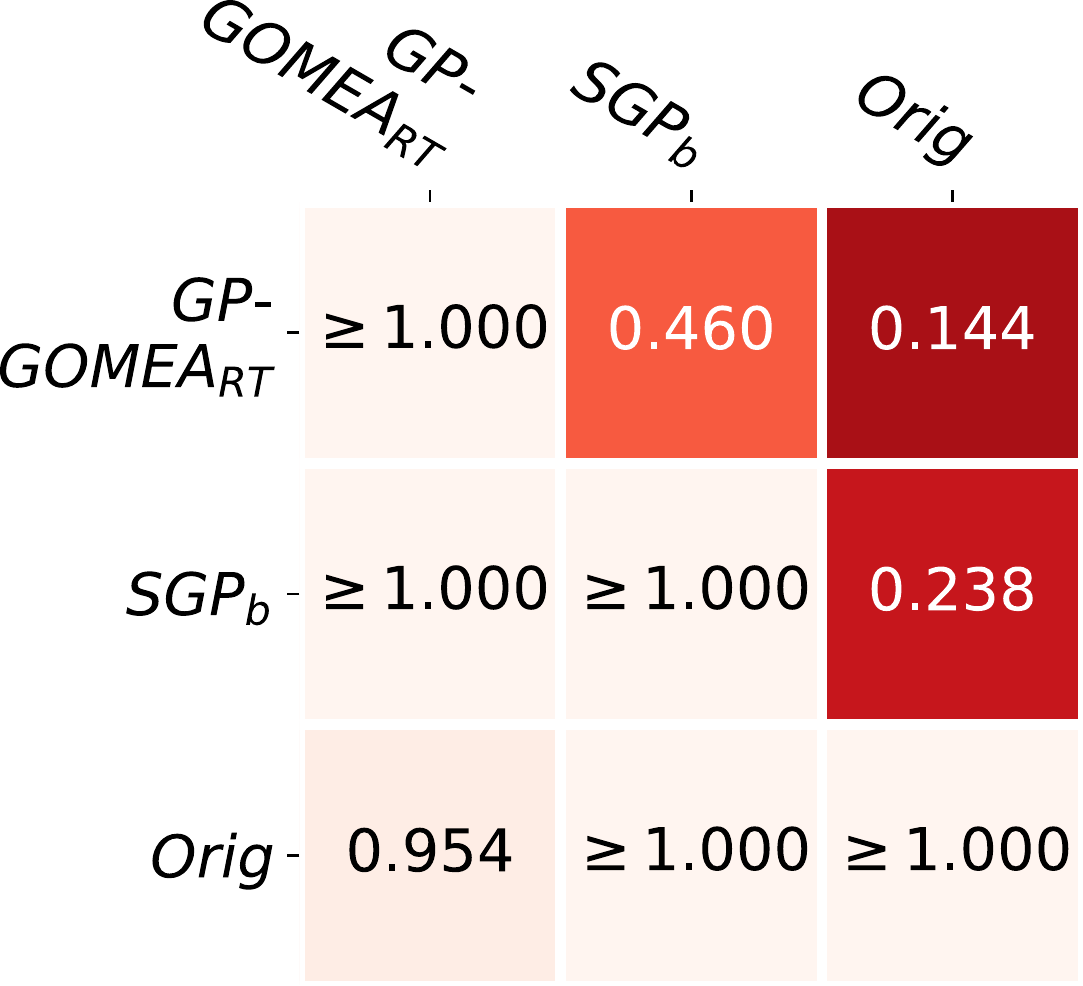} 
& 
\includegraphics[width=\linewidth]{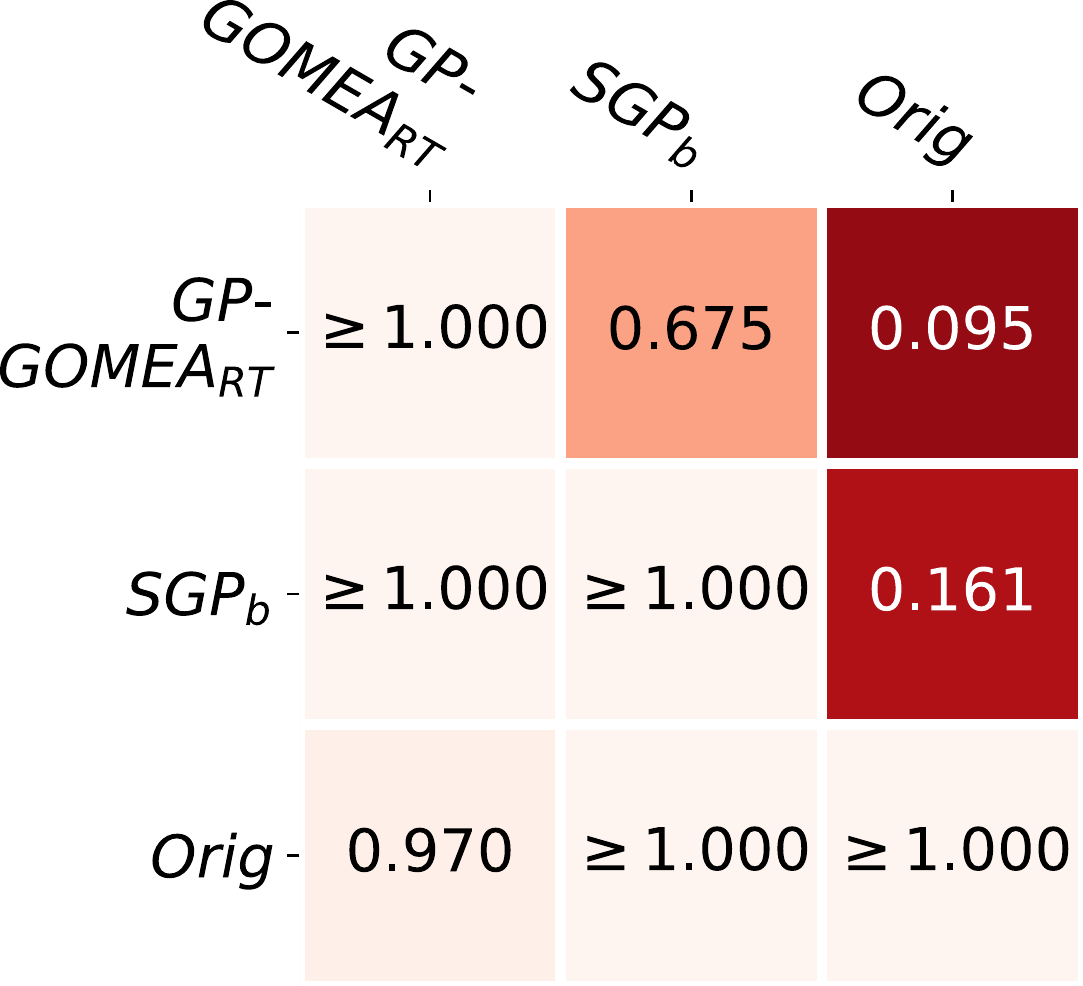}  
\\
\begin{sideways}Regression\end{sideways}
&
\includegraphics[width=\linewidth]{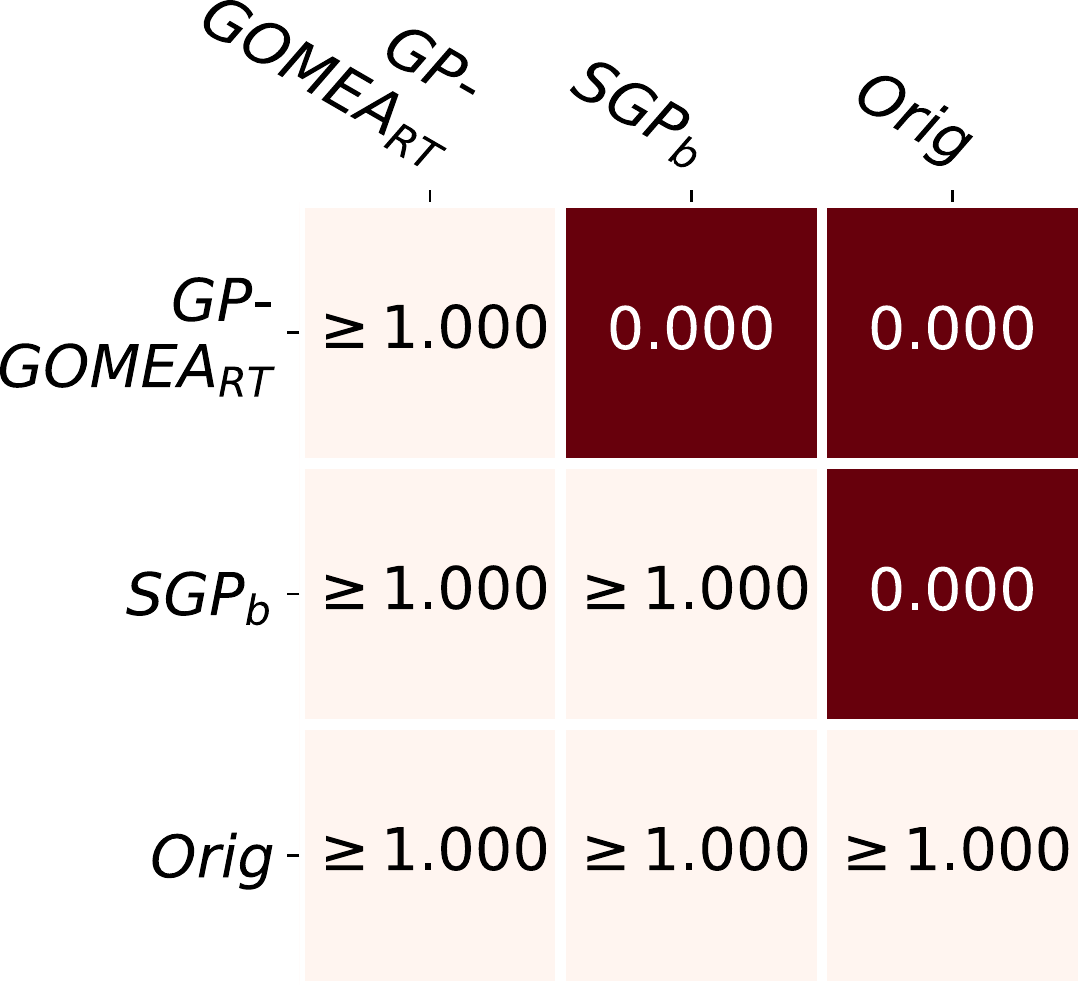} 
& 
\includegraphics[width=\linewidth]{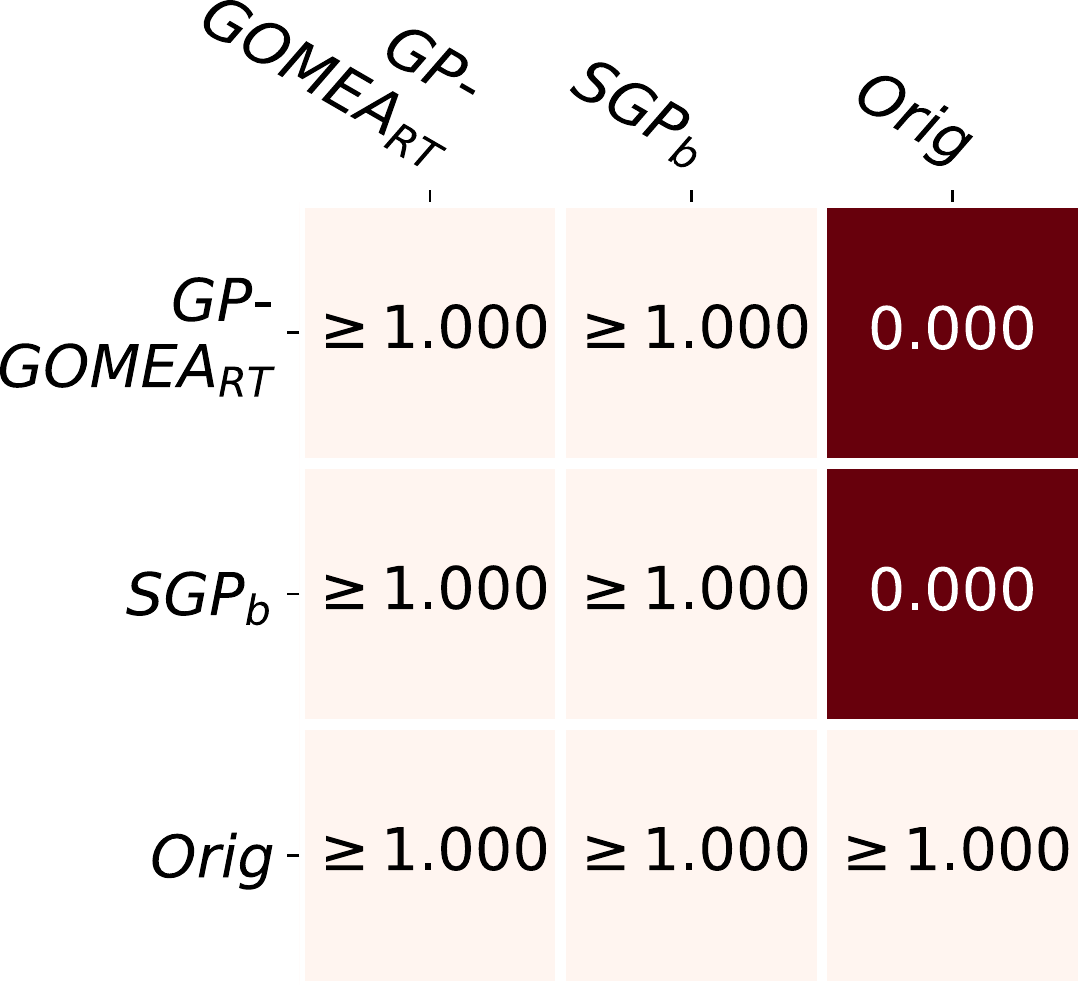}  
\\
\small
&\begin{center}$ \ \ \ \ \ h=2$ \end{center}& \begin{center}$h=4$\end{center}\\
\end{tabular}
}
\vspace{-2mm}
\caption{Holm-corrected $p$-values of pairwise Wilcoxon tests on test performance. Rows are tested to be significantly better than columns. \emph{Orig} stands for the original feature set.}\label{fig:signiftests}
\end{figure}

As for classification datasets, the Friedman test indicates that differences are presents between the treatments. Figure~\ref{fig:signiftests} (bottom) shows the Holm-corrected $p$-values obtained by the pairwise Wilcoxon tests for regression.

The statistical tests confirm the hypothesis that the algorithms are capable of providing constructed features that are more informative than the original feature set, as observed in Fig.~\ref{fig:means_median_regr} for the regression datasets. Now, GP-GOMEA\textsubscript{RT} is significantly better than SGP\textsubscript{b} when $h=2$.
For $h=4$, instead, GP-GOMEA\textsubscript{RT} is not found to be significantly better than SGP\textsubscript{b}.

\subsection{Statistical significance: two constructed features vs. the original feature set per ML algorithm}\label{sec:statanal-per-mla}
Results presented in Sec.~\ref{sec:performancefcs} indicate that our FCS brings most benefit if used with the weak ML algorithms. We now report, for each ML algorithm, on how many datasets 2 features constructed using GP-GOMEA (with $h=2$ and $h=4$) lead to statistically significantly (using Holm-corrected pairwise Wilcoxon test, $p\text{-value} < 0.05$) better, equal, or worse results compared to using the original feature set on the test set. This is shown in Table~\ref{tab:comp-mla-problems}. 

These results confirm what seen in Figures~\ref{fig:means_median_cls} and~\ref{fig:means_median_regr}. Using FCS typically outperforms the use of the original feature set for the weak ML algorithms. For the strong ML algorithms, in most cases, using the original feature set is preferable. However, for some datasets reducing the space to two compact features without compromising performance is still possible.

The use of the original feature set is generally hardest to beat when adopting RF or XGB. For RF, in the regression case with $h=4$, FCS brings benefits on the datasets Airfoil, Energy Cooling, Energy Heating, and Yacht; and performs on par with the use of the original feature set on the datasets Boston Housing and Concrete. These datasets are the ones with the smallest number of original features. We find similar results for SVM and for XGB. In the latter case, FCS is, in terms of statistical significance, equal to the original feature set only on Energy Cooling, Energy Heating, and Yacht. It is reasonable to expect that FCS works well when few features can be combined.

In the classification case, findings are different. For RF and $h=4$, the datasets where using two constructed features bring similar or better results than using the original feature set are Breast Cancer Wisconsin and Iris. The latter does have a small number of original features (4), but the former has more than several other datasets (29). Furthermore, the datasets where FCS helps are different for SVM: FCS performs equally good to the original feature set on Iris and Cylinder Bands (39 features), and better on Madelon (500 features) and Image Segmentation (19 features). Regarding XGB, there is no dataset where FCS is superior to the original feature set, but it is also not worse on almost half of the datasets.
For classification datasets, we cannot conclude that a small cardinality of the original feature set is a good indication feature construction will work well. Furthermore, feature construction influences different ML algorithms in different ways.

\begin{table}
\caption{Number of datasets where using two features constructed with GP-GOMEA results in significantly better/equal/worse test performance compared to using the original feature set.}\label{tab:comp-mla-problems}
\centering
\scalebox{0.9}{
\begin{tabular}{l c c c c c}
\toprule
& $h$ &\multicolumn{1}{c}{NB} & \multicolumn{1}{c}{SVM} & \multicolumn{1}{c}{RF} & XGB\\
\midrule
\parbox[t]{1mm}{\multirow{2}{*}{\rotatebox[origin=c]{90}{Class.}}}
& $2$ & 8/1/1 & 2/2/6 & 1/1/8 & 0/4/6 \\
& $4$ & 8/1/1 & 2/2/6 & 1/1/8 & 0/4/6 \\
\midrule
& $h$ &\multicolumn{1}{c}{LR} & \multicolumn{1}{c}{SVM} & \multicolumn{1}{c}{RF} & XGB\\
\midrule
\parbox[t]{1mm}{\multirow{2}{*}{\rotatebox[origin=c]{90}{Regr.}}}
& $2$ & 5/3/2 & 5/2/3 & 4/0/6 & 0/2/8 \\
& $4$ & 7/1/2 & 5/2/3 & 4/2/4 & 0/3/7 \\
\bottomrule
\end{tabular}
}
\end{table}

\section{Results: performance on a highly-dimensional dataset}\label{sec:gene-expr}
\begin{figure}
\centering
\includegraphics[width=0.65\linewidth]{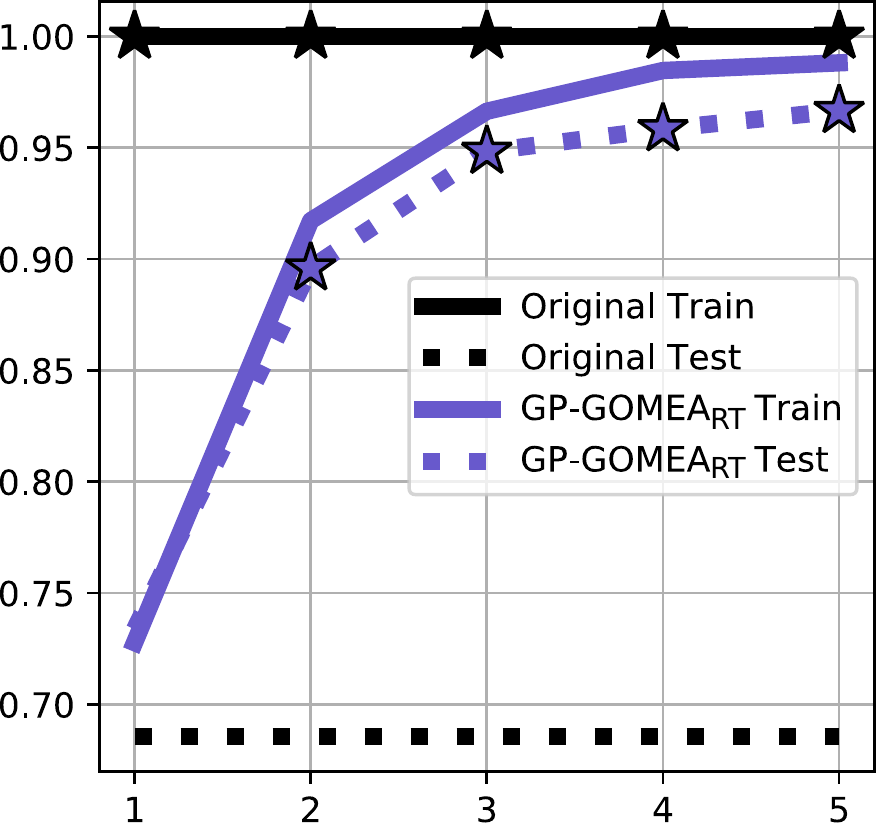}
\caption{Comparison between the use of the original feature set and FCS with GP-GOMEA\textsubscript{RT} ($h=4$) on high-dimensional gene expression data. The vertical axis reports the median F1 score, the horizontal axis reports the number of features constructed by FCS. Stars indicate statistical significant superiority ($p\text{-value} < 0.05$) of one method w.r.t. the other.}\label{fig:geneexpression}
\end{figure}

We further consider the RNA-Seq cancer gene expression dataset, comparing FCS by GP-GOMEA\textsubscript{RT} with $h=4$ against the use of the original feature set, when using NB. Fig.~\ref{fig:geneexpression} shows that NB with the original feature set overfits: the training performance is maximal, while the test performance reaches an F1 of approximately 0.65. Even tough NB is typically considered a weak estimator, the system described by the data is so severely underdetermined (over $20,000$ features vs less than $1,000$ examples) that actual patterns cannot be retrieved. The use of FCS forces NB to use only a small number of constructed features, which, in turn, can contain only a small number of the original features. Essentially, FCS provides both the advantages of feature construction and feature selection. This leads to large F1 scores already when solely two features are constructed.

\section{Results: improving interpretability}\label{sec:result-rq3}
The results presented in Sec.~\ref{sec:result-rq12} and~\ref{sec:gene-expr} showed that the original feature set can be already outperformed by two small constructed features in many cases. We now aim at assessing whether constraining features size can enable interpretability of the features themselves, as well as if extra insight can be achieved by plotting and visualizing the behavior of a trained ML model in the new two-dimensional space.

\subsection{Interpretability of small features}\label{sec:interpretability}
Table~\ref{tab:features-examples} shows some examples of features constructed by GP-GOMEA\textsubscript{RT}, for $h=2$ and $h=4$. We report the first feature constructed for the $K=2$ case, with median test performance. We show the first feature as it is typically not smaller than the second (see Fig.~\ref{fig:size-all}). Analytic quotients and protected logarithms are replaced by their respective definitions. We remark that we do not check whether the meaning of the features is sound (e.g., ensuring a certain unit of measure is returned). Constraining feature meaning is problem-dependent, and outside the scope of this work.

For classification, we choose NB as it is the method which benefits most from feature construction. The dataset considered is Ecoli, where NB achieves the largest median test improvement when $K=2$: from $F1=0.51$ with the original set, to $F1=0.63$ for $h=2$, and to $F1=0.66$ for $h=4$.

For regression, we consider LR on the Concrete dataset, for the same aforementioned reasons. The test $R^2$ obtained with the original feature set is $0.59$, the one with two features constructed by GP-GOMEA\textsubscript{RT} is $0.76$ ($0.78$) for $h=2$ ($h=4$).

For $h=2$, we argue that constructed features are mostly easy to interpret. For example, the feature shown for LR on Concrete tells us that aging ($x^{(8)}$) has a negative impact on concrete compressive strength, whereas using more water ($x^{(4)}$) than cement ($x^{(1)}$) has a positive effect (both features are in $\text{kg/cm}^3$). The impact of other features is less important (within the data variability of the dataset). 
For $h=4$, some features can be harder to read and understand, however many are still accessible. This is mostly because, even though the total solution size reachable with $h=4$ is 31, constructed features are typically half the size (see Fig.~\ref{fig:size-all}).

The features constructed for the RNA-Seq gene-expression dataset by GP-GOMEA\textsubscript{RT} in Sec.~\ref{sec:gene-expr} are also not excessively complex to be understood. For example, the first two features for the median run are:
\begin{align*}
& 1\text{st}: \sqrt{ \left(x^{(18382)}\right)^2+x^{(8014)}+x^{(3885)}+x^{(17316)}}
\\
& 2\text{nd}: \left(x^{(7491)}+\sqrt{x^{(7296)}}+x^{(19333)}\right) \times\\   & {\tiny  \left( \frac{x^{(5524)}+x^{(18053)}}{\sqrt{1 +  \left(x^{(5579)}-x^{(4417)}\right)^2}}+x^{(14153)}+x^{(19751)}- \frac{ x^{(13744)} }{\sqrt{1 + \left( x^{(16581)} \right)^2 } }\right) }%
\end{align*}
Even tough the second feature is somewhat involved, it is arguably still possible to carefully analyze it and obtain a picture of how gene expression levels interact.

Overall, we cannot draw a strict conclusion on whether the features found by our approach are interpretable, as interpretability is a subjective matter and, to date, no clear-cut metric exists~\cite{lipton2018mythos,guidotti2018survey} (we discuss this more in Sec.~\ref{sec:discussion}). Yet, it appears evident that enforcing a restriction on their size is a necessary condition. We generally find that features using 15 or more nodes start to be hard to interpret w.r.t. our experimental settings, i.e., using our function set.
Lastly, features constructed without a strict size limitation (by SGP) are in general very large, and thus extremely hard to understand. As an example, Figure~\ref{fig:sgp-feature} shows the first of the two features with median test performance constructed by SGP for LR on Concrete (this is smaller than the first feature found by SGP for NB on Ecoli).

\begin{table}
\caption{Examples of features constructed by GP-GOMEA\textsubscript{RT} with $h \in \{2,4\}$, $K=2$, for NB on Ecoli, and for LR on Concrete.}
\label{tab:features-examples}
\small
\centering
\scalebox{0.9}{
\begin{tabular}{l|c|c}
\toprule
 & $h$ & 1st Feature \\
 \midrule
 \parbox[t]{2mm}{\multirow{3}{*}{\rotatebox[origin=c]{90}{NB}}}
 & 2 & 
 $ x^{(3)} + x^{(6)} + x^{(1)} / \sqrt{ 1 + \left( x^{(6)} \right)^2 }$
 \\   \cline{2-3}  
 & 4 &
 $ x^{(6)} \left(x^{(7)}\right)^2 x^{(3)} + 0.144 / \sqrt{1 + \left( \exp(x^{(2)}) \right)^2} - x^{(1)}x^{(2)}x^{(5)}$ \\ 
\midrule
 \parbox[t]{2mm}{\multirow{2}{*}{\rotatebox[origin=c]{90}{LR}}} 
 & 2 &
 $x^{(4)} - x^{(1)} + 932.204 / \sqrt{ 1 + (x^{(8)})^2 } $ 
 \\
 \cline{2-3}
 & 4 & 
$ \sqrt{ 19.764 \log | x^{(8)} | } + x^{(2)} + 2x^{(1)} / \sqrt{1 + (x^{(4)})^2}$\\
\bottomrule
\end{tabular}
}
\end{table}

\begin{figure}
\centering
\scalebox{0.68}{
\parbox{1.0\linewidth}{
\begin{align*}
\biggl(&
\log_p(((((((x^{(4)}+x^{(2)})\div x^{(4)}) \times (x^{(1)} \div x^{(4)} \div x^{(4)} \times ( x^{(1)}\div x^{(4)} x^{(8)} 1065.162 \times \\
& \log_p((((x^{(4)}x^{(1)}+x^{(1)}\div x^{(8)}\div x^{(8)}) \times (x^{(4)}-(x^{(1)}+ ((x^{(2)} \div x^{(4)}+ \log_p (x^{(1)})))^2)))+\\
&-(x^{(5)}x^{(1)}+x^{(2)} \div x^{(8)}+(((x^{(8)}+x^{(6)}+x^{(2)}) \div ((x^{(1)} \div x^{(4)})+ \sqrt{x^{(4)}} \div x^{(2)} ))+ \\
&\exp((((x^{(8)}+x^{(2)}) \div x^{(8)}+(x^{(4)}+x^{(2)}) \div x^{(4)}) \div x^{(5)}+x^{(6)} \div x^{(4)}))))))))) \div x^{(7)}) \times\\
& (x^{(8)}-(441.237+x^{(2)}))) -x^{(1)})) \biggr)^\frac{1}{2}
\end{align*}
}
}
\caption{Example of a relatively ``small'' feature constructed by SGP, derived from a tree with 96 nodes. 
Note that the analytic quotient operator ($\div$) and the protected logarithm ($\log_p$) are \emph{not} expanded to their respective definitions to keep the feature contained.
This feature is arguably very hard to interpret.}\label{fig:sgp-feature}
\end{figure}

\subsection{Visualizing what the ML algorithm learns}
The construction of a small number of interpretable features can enable a better understanding of the problem and of the learned ML models. The case where up to two features are constructed is particularly interesting, since it allows visualization.

We provide one example of classification boundaries and one of a regressed surface, inferred by SVM on a two dimensional feature space obtained with our approach using GP-GOMEA\textsubscript{RT}. 

\begin{figure}
\centering
\includegraphics[width=0.89\linewidth]{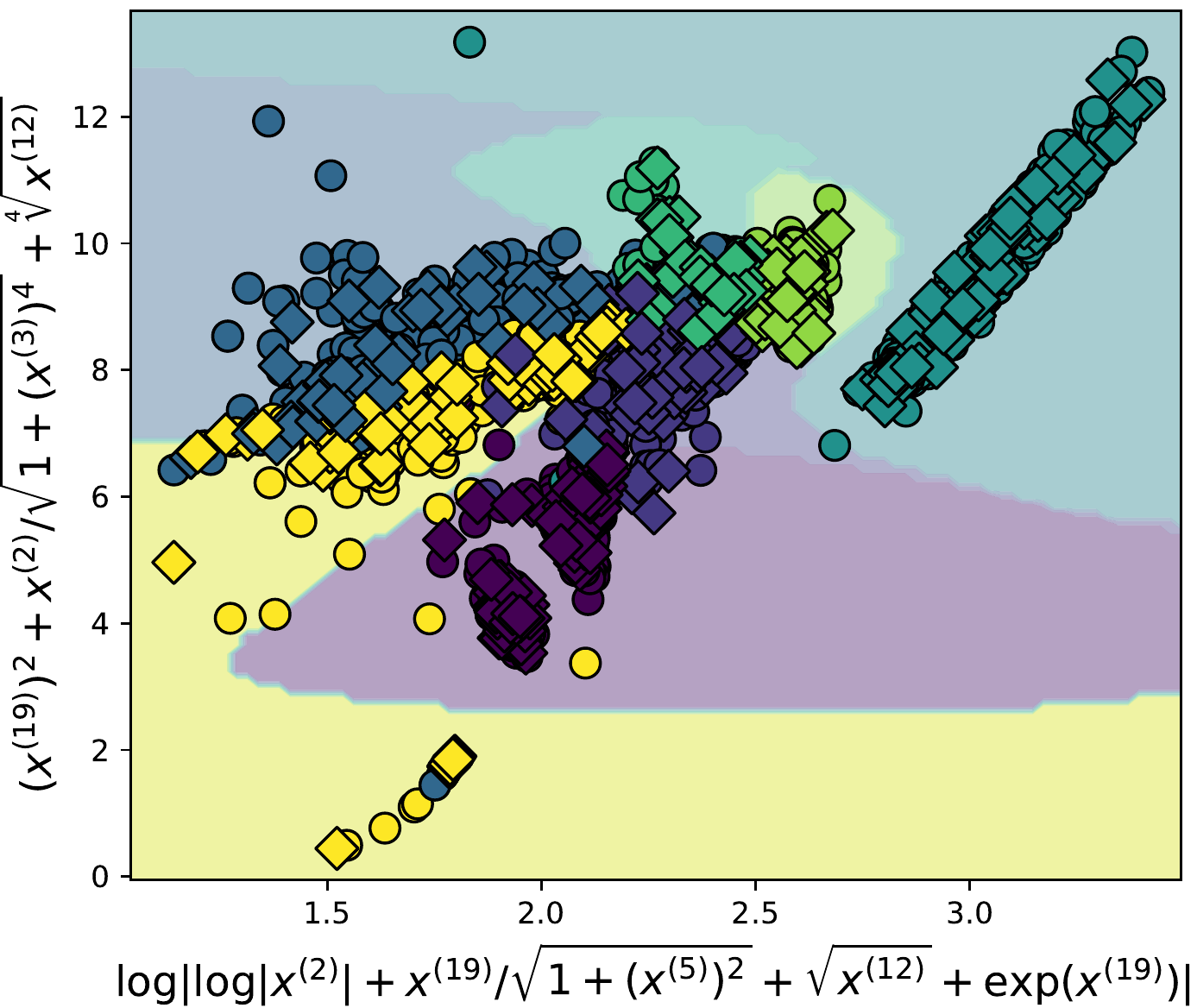}
\caption{Classification boundaries learned by SVM with two features constructed by GP-GOMEA\textsubscript{RT} ($h=4$) on the Image Segmentation dataset. The run with median test performance is shown. Circles are training samples, diamonds are test samples.}\label{fig:svm-segmentation}
\end{figure}

The classification dataset on which we find the best test improvement for $h=4$ is Image Segmentation, where the $F1$ score of SVM reaches 0.88, against 0.65 using the original feature set (median run).
Figure~\ref{fig:svm-segmentation} shows the classification boundaries learned by SVM. The analytic quotient operator $\div$ and the protected log $\log_p$ are replaced by their definition for readability.
The constructed features are rather complex here, yet readable. At the same time, it can be clearly seen how the training and test examples are distributed in the 2D space, and what classification boundaries SVM learned.

For regression, Figure~\ref{fig:svm-yacht} shows the surface learned by SVM on Yacht (median run), where GP-GOMEA\textsubscript{RT} with $h=2$ constructs two features that lead to an $R^2$ of 0.98, against 0.85 obtained using the original feature set. The features are arguably easy to interpret, while it can be seen that the learned surface accurately models most of the data points.

\section{Running time}\label{sec:running-time}
Our results are made possible by evaluating the fitness of constructed features with cross-validation, a procedure which is particularly expensive.
Table~\ref{tab:running-time} shows the (mean over 30 runs) serial running time to construct five features on the smallest and largest classification and regression datasets, using GP-GOMEA\textsubscript{RT} with $h=4$ and the parameter settings of Sec.~\ref{sec:experiments}, on the relatively old AMD Opteron\texttrademark~Processor 6386 SE\footnote{http://cpuboss.com/cpu/AMD-Opteron-6386-SE}.
Running time has a large variability, from seconds to dozens of hours, depending on dataset size and ML algorithm. For the traditional datasets and ML algorithms we considered, it can be argued that our approach can be used in practice. However, for very high-dimensional datasets, only fast ML algorithms can be used. The construction of 5 features for the RNA-Seq gene expression dataset took 25 minutes even tough NB was used. To use slower ML algorithms would easily require dozens to hundreds of hours. 

As to memory occupation, it basically mostly depends on the way the chosen ML algorithm handles the dataset. Our runs required at most few hundreds of MBs when dealing with the larger traditional datasets, for SVM and RF. Handling the parallel execution of FCS experiments upon the gene expression dataset required a few GBs.

\begin{table}
\caption{Mean serial running time to construct five features using GP-GOMEA\textsubscript{RT} ($h=4$) on the smallest and largest traditional datasets.}
\label{tab:running-time}
\small
\centering
\scalebox{0.80}{
\begin{tabular}{c|l|c|c|c|c|c}
\toprule
 & Dataset & Size & NB/LR & SVM & RF & XGB\\
 \midrule
 \parbox[t]{2mm}{\multirow{2}{*}{\rotatebox[origin=c]{90}{Clas.}}} 
 & Iris & $150\times4$ & $7$ s & $ 2$ m  & $25$ m & $42$ m\\
 & Madelon & $2600\times500$ & $ 4$ m  & $14$ h & $8$ h & $10$ h\\
 \midrule
 \parbox[t]{2mm}{\multirow{2}{*}{\rotatebox[origin=c]{90}{Regr.}}} 
 & Yacht & $308\times7$ & $8$ s & $4$ m & $1$ h & $1$ h\\
 & Tower & $4999\times26$ & $2$ m& $34$ h & $34$ h & $13$ h\\
\bottomrule
\end{tabular}
}
\end{table}

\section{Discussion}\label{sec:discussion}
We believe this is one of the few works on evolutionary feature construction where the focus is put on both improving the performance of an ML algorithm, and on human interpretability at the same time. The interpretability we aimed for is twofold: understanding the meaning of the features themselves, as well as reducing their number. GP algorithms are key, as they can provide constructed features as interpretable expressions given basic functional components, and a complexity limit (e.g., tree height).

We have run a large set of experiments, totaling more than 150,000 cpu-hours. Our results strongly support the hypothesis that the original feature set can be replaced by few (even solely $K=2$) features built with our FCS without compromising performance in many cases. In some cases, performance even improved. GP-GOMEA\textsubscript{RT} and SGP\textsubscript{b} achieve this result while keeping the constructed feature size extremely limited ($h=2,4$). SGP leads to slightly better performance than GP-GOMEA\textsubscript{RT} and SGP\textsubscript{b}, but at the cost of constructing five to ten times larger features. RS proved to be less effective than the GP algorithms.

Our FCS is arguably most sensible to use for simpler ML algorithms, such as NB and LR. Constructed features change the space upon which the ML algorithm operates. SVM already includes the kernel trick to change the feature space. Similarly, the trees of RF and XGB effectively embody complex non-linear feature combinations to explain the variance in the data. NB and LR, instead, do not include such mechanisms. Rather, they have particular assumptions on how the features should be combined (NB assumes normality, LR linearity). The features constructed by GP can transform the input the ML algorithm operates upon, to better fit its assumptions.

We found that performance was almost always significantly better than compared to using the original feature set for NB and LR. As running times for these ML algorithms can be in the order of seconds or minutes (Sec.~\ref{sec:running-time}), feature construction has the potential to be routinely used in data analysis and machine learning practice. Furthermore, FCS (or a modification where the constructed features are added to the original set) can be used as an alternative way to tune simple ML algorithms which have limited or no hyper-parameters.

We have shown that our approach can also be helpful when dealing with high-dimensional data (on the RNA-Seq gene expression dataset), where system underdetermination can cause even simpler ML algorithms to overfit. This is because FCS essentially embodies feature selection, as we only construct a small number of small-sized features. 

We remark that we did not adopt very popular high-dimensional datasets concerning image recognition such as MNIST~\cite{lecun1998mnist}, CIFAR~\cite{krizhevsky2009learning}, or ImageNet~\cite{deng2009imagenet}. In these datasets, features represent pixels, and each pixel has no particular meaning. Consequently, constructing features as readable pixel transformations will likely carry no unhelpful information to explain the behavior of a ML model.

Regarding the comparison between the search algorithms, GP-GOMEA\textsubscript{RT} was found to be slightly preferable to SGP\textsubscript{b} (especially for $h=2,K=2$). 
We believe that significantly better results can be achieved if bigger population sizes and larger evaluations budgets can be employed (we kept the population size limited due to the computational expensiveness of SVM and RF). 

\begin{figure}
\centering
\tabcolsep=0.00mm
\begin{tabular}{cc}
\includegraphics[width=0.48\linewidth]{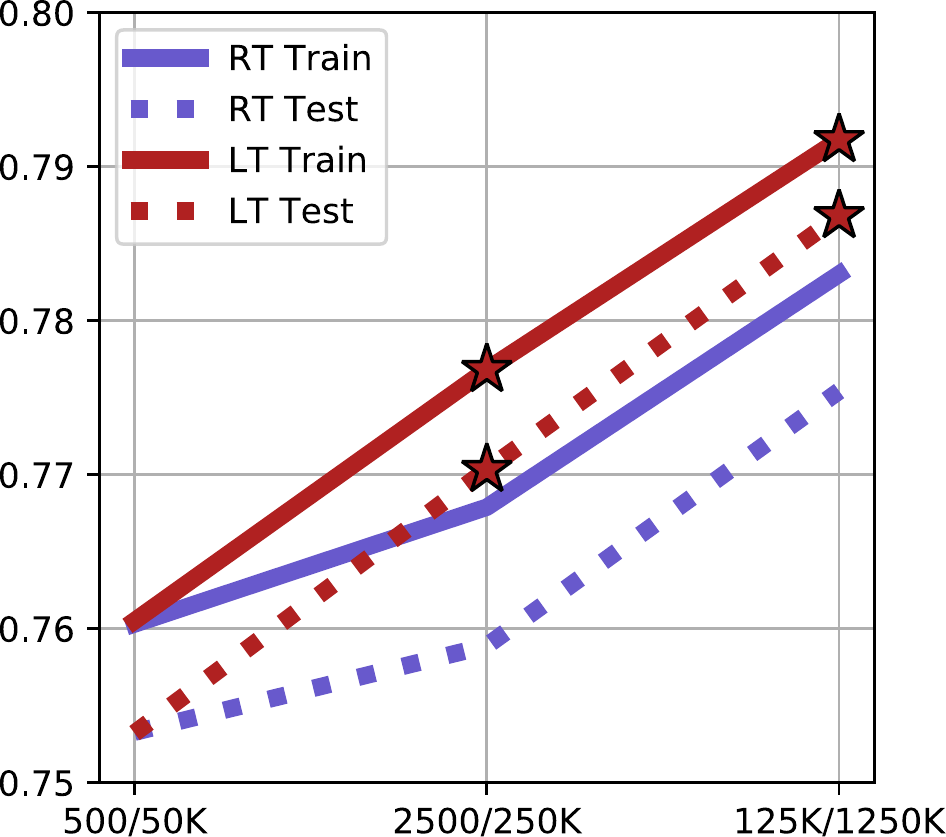}&
\includegraphics[width=0.48\linewidth]{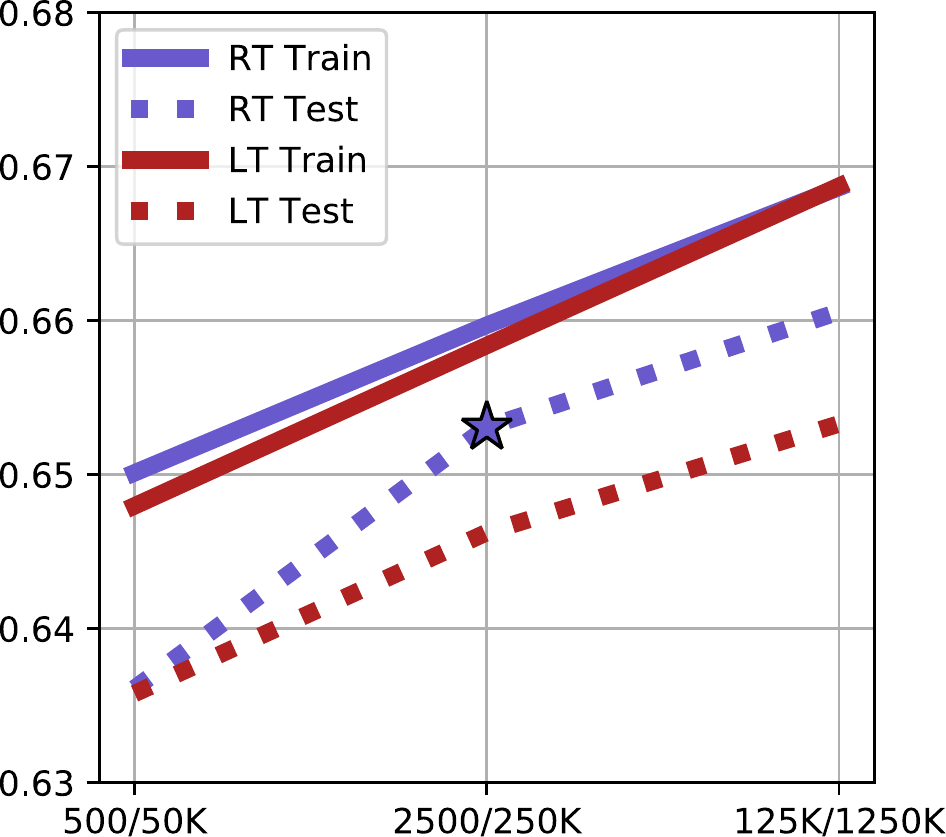}\\
\end{tabular}
\caption{Comparison between the use of the RT and of the LT in GP-GOMEA. Vertical axis: median F1 score of 30 runs, obtained by NB on Image Segmentation (left) and on Madelon (right) using the first constructed feature, with $h=4$ (note the different scale). Horizontal axis: population size / fitness evaluations budget. Stars indicate significant superiority ($p-\text{value} < 0.05$) of one method w.r.t. the other.}\label{fig:fos-comparison}
\end{figure}

Particularly for GP-GOMEA, previous work has shown that having sufficiently large population sizes enables the possibility to exploit linkage estimation and perform better-than-random mixing~\cite{virgolin2019model,virgolin2017scalable}. 
To validate this also within the framework of our proposed FCS, we scaled the population size and the budget of fitness evaluations, and compared the use of the LT with the use of the RT, on two traditional classification dataset: Image Segmentation (19 features) and Madelon (500 features), using NB. The outcome is shown in Figure~\ref{fig:fos-comparison}: the employment of big-enough population sizes (and of sufficient numbers of fitness evaluation) can lead to better performance, \emph{if statistical metrics can be measured reliably}. For Image Segmentation, the number of terminals to be considered in the genotype is relatively small due to the use of 19 features. This allows the LT to estimate node interdependencies reliably, and deliver better-than-random performance. For Madelon, the large number of terminals (500 features) makes it hard for the LT to outperform the RT within a limited computational budget.
All in all, we recommend the use of GP-GOMEA as feature constructor since it was not worse on classification and was statistically better for regression. Furthermore, we advice to use the LT if the population size can be of the order of thousands or more (or even better, if exponential population sizing schemes are used as in~\cite{virgolin2019model,virgolin2017scalable}). Otherwise, the RT should be preferred.

To assess if small constructed features are interpretable and if it is possible to visualize what the behavior of learned ML models, we showed some examples, providing evidence that both requirements can be reasonably satisfied. However, we did not perform a thorough study on interpretability of the constructed features. Several metrics have been recently proposed to measure some form of interpretability for ML models, that could be used to measure the interpretability of features as well. E.g., in~\cite{lipton2018mythos} two metrics called \emph{simulatability} and \emph{decomposability} are proposed. Simulatability represents the capability of humans to predict the output of a model given an input. Decomposability represents the capacity to intuitively understand the components of a model. Crucially, to measure this type of metrics, user studies need to be conducted. For example, experts of a field should be asked to provide feedback, on features constructed for datasets they are knowledgeable about (e.g., biochemists for data on gene expression, civil engineers for data on concrete strength). Nonetheless, we believe that enforcing features (and GP programs in general) to be small still remains a necessary condition to allow interpretability, although it is often ignored in GP literature~\cite{virgolin2019model}.

Considering the visualization examples proposed in Section~\ref{sec:result-rq3}, it is natural to compare our approach with well-known dimensionality reduction techniques, such as Principal Component Analysis (PCA)~\cite{wold1987principal} or t-Distributed Stochastic Neighbor Embedding (t-SNE)~\cite{maaten2008visualizing}. We remark that those techniques and our FCS have very different objectives. In general, the sole aim of such techniques is to reduce the data dimensionality. PCA does so by detecting components that capture maximal variance. However, it does not attempt to optimize the transformation of the original feature set to improve an ML algorithm's performance. Also, PCA does not focus on the interpretability of the feature transformations. FCS takes the performance of the ML algorithm and interpretability of the features into account, while dimensionality reduction comes from forcing the construction of few features. We compared using 2 features constructed with RS (the worst search algorithm) with maximum $h=2$, with using the first 2 PCs found by PCA. The use of constructed features over PCs resulted in significantly superior or equal test performance for all ML algorithms and for all problems. We remark, however, that PCA is extremely fast and independent from the ML algorithm.

Our FCS has several limitations. A first limitation regards the performance obtainable by the ML algorithm using the constructed features. FCS is iterative, and this can lead to suboptimal performance for a chosen $K$, compared to attempting to find $K$ features at once.
This is because the contributions of multiple features to an ML algorithm are not necessarily perpendicular to each other~\cite{tran2019genetic}.
FCS could be changed to find at any given iteration, a synergistic set of $K$ features, that is independent from previous iterations. 
To this end, larger population sizes need to be employed, and the search algorithms need to be modified so that they can evolve sets of constructed features (a similar proposal for SGP was done in~\cite{krawiec2002genetic}). 
Yet, it is reasonable to expect that if $K$ features need to be learned at the same time, larger population sizes may be needed compared to learning the $K$ features iteratively.

Another limitation of this work is that hyper-parameter tuning was not considered. To include hyper-parameter tuning within FCS could bring even higher performance scores, or help prevent overfitting. 
A possibility could be, for example, to evolve pairs of features and hyper-parameter settings, where every time a feature is evaluated, the optimal hyper-parameters are also searched for. Such a procedure may likely require strong parallelization efforts, as $C$-fold cross-validation should be carried out for each combination of hyper-parameter values.

Lastly, it would be interesting to extend our approach to other classification and regression settings, e.g., problems with missing data; or to unsupervised tasks, as simple features may lead to better clustering of the examples.

\section{Conclusion}\label{sec:conclusion}
With a simple evolutionary feature construction framework we have studied the feasibility of constructing few crucial and compact features with Genetic Programming (GP), towards improving the explainability of Machine Learning (ML) models without losing prediction accuracy.
Within the proposed framework, we compared standard GP, random search, and the GP adaptation of the Gene-pool Optimal Mixing Evolutionary Algorithm (GP-GOMEA) as feature constructors, and found that GP-GOMEA is overall preferable when strict limitations on feature size are enforced. 
Despite limitations on feature size, and despite the reduction of problem dimensionality that we imposed by constructing only two features, we obtained equal or better ML prediction performance compared to using the original feature set for more than half the combinations of datasets and ML algorithms.
In many cases, humans can understand what the feature means, and it is possible to visualize how trained ML models will behave. 
All in all, we conclude that feature construction is most useful and sensible for simpler ML algorithms, where more resources can be used for evolution (e.g., larger population sizes), which, in turn, unlock the added benefits of more advanced evolutionary mechanisms (e.g., using linkage learning in GP-GOMEA). 

\section*{Acknowledgments}
The authors acknowledge the Kinderen Kankervrij foundation for financial support (project \#187). The majority of the computations for this work were performed on the Lisa Compute Cluster with the support of SURFsara.

\bibliographystyle{elsarticle-num}
\bibliography{featureconstructiongomea} 
\end{document}